\PassOptionsToPackage{table}{xcolor}
\documentclass[letterpaper]{article} % DO NOT CHANGE THIS
\usepackage[preprint]{aaai2027}
\usepackage[hyphens]{url}  % DO NOT CHANGE THIS
\usepackage{graphicx} % DO NOT CHANGE THIS
\usepackage{natbib}  % DO NOT CHANGE THIS AND DO NOT ADD ANY OPTIONS TO IT
\usepackage{caption} % DO NOT CHANGE THIS AND DO NOT ADD ANY OPTIONS TO IT
\usepackage{algorithm}
\usepackage{algorithmic}

\usepackage{newfloat}
\usepackage{listings}
\DeclareCaptionStyle{ruled}{labelfont=normalfont,labelsep=colon,strut=off} % DO NOT CHANGE THIS
\floatstyle{ruled}
\newfloat{listing}{tb}{lst}{}
\floatname{listing}{Listing}

\usepackage{booktabs}
\usepackage{color}
\newcommand{\tzlconfirmed}[1]{{#1}}
\usepackage{subcaption}
\usepackage{amsmath}
\usepackage{amssymb}
\usepackage{multirow}
\usepackage{siunitx}
\usepackage[table]{xcolor}
\usepackage{adjustbox}
\usepackage{array}
\usepackage{tabularx}
\usepackage{makecell}
\usepackage{cuted}
\usepackage[most]{tcolorbox}
\usepackage{enumitem}
\newtcolorbox{riskbox}{
    enhanced,
    colback=white,
    colframe=black!55,
    boxrule=0.45pt,
    arc=1pt,
    left=5pt,
    right=5pt,
    top=4pt,
    bottom=4pt,
    before skip=0pt,
    after skip=0pt
}
\definecolor{oursgray}{gray}{0.92}

\title{Tail-risk-aware Agent Safety: Constraining Tail Risks \\ Across Multi-step LLM Agent Workflows}
\author{
    Shilong Pan\textsuperscript{\rm 1},
    Zhiliang Tian\textsuperscript{\rm 1}\corresponding,
    Zihan Chen\textsuperscript{\rm 1},
    Wentao Zou\textsuperscript{\rm 1},
    Weiguo Chen\textsuperscript{\rm 1},
    DongSheng Li\textsuperscript{\rm 1}\corresponding,
}
\affiliations{
    \textsuperscript{\rm 1}PDL Lab, College of Computer Science and Technology\\
    National University of Defense Technology
    panshilong18@nudt.edu.cn
}

\begin{document}

\maketitle

\begin{abstract}
% \begin{abstract}
LLM-based agents extend LLMs with tool calls, memory, planning, and retrieval to act in external environments, but they are vulnerable to attacks at different workflow steps.
Existing defenses, including prompt engineering and post-training alignment, primarily improve LLM-only safety and often overlook risks that arise from multi-step execution.
% Existing defenses, including prompt engineering and post-training alignment, typically rely on handcrafted safety templates or training LLMs with safety-related datasets and reward mechanisms to refuse harmful queries. 
% They focus on LLM-only safety while overlooking risks that arise from multi-step execution.
External verification adds guardrails to check each step separately, but it typically ignores how risks can propagate across steps.
% External verification strategies introduce external guardrails to check the safety of each step independently along the workflow, while ignoring the interactive impact (e.g., risk transitions) across steps through the workflow.
We argue that we should consider risk transitions across dependent steps through the agent workflow, rather than separate verification on a single step.
In this paper, we propose \textbf{T}ail-risk-aware \textbf{A}gent \textbf{S}afety (TAS), a tail-risk-aware safety framework that models agent workflow safety as a sequence of dependent safety states and constrains the rare but extreme risks (i.e., tail risk) with a shared budget.
TAS first models safety-state transitions sequentially along the workflow as an MDP to trace risk propagation, then measures a tail-sensitive risk to target rare but high-cost risks (i.e., tail risk), and constrains the risk with a shared budget to derive a step-wise safety strategy.
% 
% formalizes the sequential agent safety state along the workflow as a Markov Decision Process (MDP) to trace risk transitions, 
% 
% then applies a tail-sensitive constraint to target rare but high-cost risks (i.e., tail risk) under an overall shared budget, and finally optimizes the constrained objective for a step-wise safety strategy.
% 
By constraining tail risk rather than the overall average, TAS suppresses rare high-cost failures while preserving utility in normal cases.
% prevents catastrophic risks while preserving utility in normal cases.
% XXX explicitly constrains tail risks from hiding in the normal overall average, which ensures the workflow remains safe in critical scenarios and maintains utility in normal scenarios.
Experiments show that TAS achieves better performance compared with eleven baselines across three benchmarks.
% six LLM backbones show that our method outperforms all baselines.
% \end{abstract}

\end{abstract}

% Uncomment the following to link to your code, datasets, an extended version or similar.
% You must keep this block between (not within) the abstract and the main body of the paper.
% Make sure that you do not de-anonymize yourself with these links.
% \begin{links}
%     \link{Code}{https://aaai.org/example/code}
%     \link{Datasets}{https://aaai.org/example/datasets}
%     \link{Extended version}{https://aaai.org/example/extended-version}
% \end{links}

\section{Introduction}
% LLM-based agents, which combine LLMs with tools, memory, and Retrieval-Augmented Generation (RAG) to interact with the external environments, \tzlconfirmed{are widely adopted in various fields \cite{openagi, xu2025mem}.
% 
% The safety of Large Language Models (LLMs) has become a critical and common concern \cite{wei2023jailbroken,zhou2024alignment}.
% 
% As LLM-based agents are increasingly adopted in various fields \cite{openagi, xu2025mem}, their safety has become a critical concern \cite{chen2024agentpoisonredteamingllmagents,yuan2024rjudge}.
% By combining LLMs with tools, memory, and retrieval to interact with external environments, agents are susceptible to diverse adversarial attacks \cite{yu2023finmem,hua2024trustagent}.
% 
\tzlconfirmed{LLM safety has drawn increasing attention \cite{wei2023jailbroken,zhou2024alignment}. 
Building on LLMs, agentic LLM systems, in which LLMs are augmented with external capabilities such as tool calling, memory, planning, and retrieval to interact with external environments \cite{openagi, xu2025mem}, are referred to as LLM-based agents in prior works \cite{qiu2024llm,zhu2025oagents}.
% Building on LLMs, tool-calling LLM systems (often referred to as LLM-based agents) augment an LLM with external capabilities such as tool use, memory, planning, and retrieval, to interact with external environments \cite{openagi, xu2025mem}.
% These agents have been widely adopted across domains .
However, the safety of these agent workflows has become a critical concern \cite{chen2024agentpoisonredteamingllmagents, yuan2024rjudge}, as they expose a wider range of possible attacks from various tool callings \cite{yu2023finmem, hua2024trustagent}.}
% 
% However, agents are susceptible to diverse adversarial attacks, raising critical safety concerns that} impede their reliable adoption in sensitive domains \cite{yu2023finmem, abbasian2024conversationalhealthagentspersonalized, hua2024trustagent}.
% 
\tzlconfirmed{Attackers can target different steps of an agent, including input, retrieval, planning, and execution, causing task failure, malicious content, or harmful actions \cite{zhan-etal-2024-injecagent, debenedetti2024agentdojo}.}
% 
% To mitigate agent safety concerns, researchers have proposed various safety mechanisms, typically by strengthening the core LLM safety or by adding external verifications to monitor the safety of an agent workflow.
% 
\tzlconfirmed{To mitigate LLM-based agent safety concerns, prior works mainly focused on constraining agent behavior to refuse unsafe queries.}
% aligning LLMs with human values to refuse unsafe queries.

\tzlconfirmed{Existing methods to improve agent safety mainly fall into two categories: prompt engineering and post-training alignment.
Prompt engineering wraps inputs with pre-designed templates for corresponding tasks during LLMs inference \cite{xie2023defending,li2024deepinceptionhypnotizelargelanguage}.
Prompt engineering requires different delicately designed safety templates in various scenarios and is vulnerable to adversarial prompt attacks that intentionally disable the safety prompt templates \cite{liu2024autodan,li2024deepinceptionhypnotizelargelanguage}.
Post-training alignment, typically including supervised fine-tuning (SFT) and reinforcement learning (RL), uses safety-related annotated datasets and reward mechanisms to train LLMs to refuse harmful queries \cite{bai2022training,guan2024deliberative,LiSARATRX25}.}
% 
% However, SFT with annotated datasets requires costly manual efforts to ensure high-quality data \cite{yuan2024refuse,yu2024robust}. RL requires carefully designed reward functions and may suffer reward hacking problems \cite{amodei2016concreteproblemsaisafety,pmlr-v202-gao23h,weng2024rewardhack}. Both of these post-training methods require amount of computational resources.
% 
\tzlconfirmed{However, SFT relies heavily on manual safety annotations, which are typically incomplete and may fail to cover diverse adversarial attacks \cite{yuan2024refuse,yu2024robust}.
RL requires carefully designed reward functions and can suffer from reward hacking problems, leading to unintended risk behaviors \cite{amodei2016concreteproblemsaisafety,pmlr-v202-gao23h,weng2024rewardhack}.}
\tzlconfirmed{Moreover, both prompt engineering and post-training focus solely on the safety of the LLM. Since an agent involves a multi-step workflow rather than only LLM single-step generation, ensuring LLM-level safety alone is not enough to guarantee the safety of the whole agent workflow. Consequently, these methods overlook the safety of other (i.e., non-LLM) components in the entire agent workflow, including retrieval, planning, and execution components.}

\tzlconfirmed{To address the above issues, recent studies apply external verification strategies. These approaches introduce additional guardrail agents and monitors to separately check the safety of each individual step in the entire agent workflow \cite{xiang2024guardagent,huang2024large}.
Although these methods cater to other non-LLM components, they evaluate the safety of each individual step independently, ignoring the interactive impact during the agent workflow.}
% 
% As a result, they actually solely improve the safety of individual step, but ignore the interactive safety \tzlconfirmed{impact among different agent steps.}
% 
\tzlconfirmed{As a result, they actually solely improve the safety of individual steps, but ignore inter-step risk propagation through the agent workflow.}

\tzlconfirmed{To improve the safety of an agent, rather than verifying each step individually, we argue that we should consider the internal interactive dependencies within an agent workflow.
Therefore, we should improve the safety of the entire agent workflow, considering both each step itself and its interactive impact.}
% 
% By modeling how safety states evolve across steps sequentially in the workflow under a unified risk constraint, we can exploit internal inter-step interactions to make the entire agent workflow safer.
% 
\tzlconfirmed{By modeling the workflow as a sequence of dependent steps, and constraining the overall risk with a single shared budget across the agent instead of independent constraints for each step, we can account for inter-step risk propagation and make the entire workflow safer.}

In this paper, we propose \textbf{T}ail-risk-aware \textbf{A}gent \textbf{S}afety (TAS), a tail-risk-aware safety framework for LLM-based agent workflows. TAS models workflow safety as a sequence of dependent safety states and constrains rare but extreme risks with a shared budget, suppressing tail risks while preserving utility in normal cases.
Specifically, the safety state at each workflow step provides a compact risk-relevant summary of the current workflow context, including the user query, intermediate outputs, and safety evaluations.
We formulate the sequential workflow as a Markov Decision Process (MDP), where actions and intermediate outputs are conditioned on the current safety state, enabling TAS to capture risk propagation along the workflow instead of checking isolated steps independently.
Furthermore, TAS introduces a tail-sensitive safety constraint to prioritize worst-case scenarios over overall average costs. Since most agent behaviors are typically safe while a small fraction may incur severe failures, an overall-average risk measure can mask rare but severe risks. Using an average-based constraint to cover these tail risks may also over-penalize normal cases and compromise utility. TAS instead constrains tail risks with a shared budget, reducing high-risk failures in critical scenarios while maintaining utility in normal cases.
Extensive experiments show that TAS improves safety while preserving utility compared to eleven defense baselines across three benchmarks.

% \tzlconfirmed{Our main contributions are threefold as follows:
% (1) We propose TAS that models internal interactions among sequential steps within an agent as a CMDP to evaluate and ensure the holistic safety.
% (2) We introduce a tail-risk-aware CMDP algorithm that explicitly constrains the worst-case scenarios while maintaining utility performance in normal scenarios.
% (3) Extensive experiments demonstrate that TAS achieves better performance compared with eleven defense baselines on three benchmarks.}

Our main contributions are threefold as follows:
(1) We propose TAS, a tail-risk-aware safety framework that models risk propagation across dependent steps in LLM-based agent workflows.
(2) We introduce a shared-budget safety strategy that prioritizes rare high-risk failures while preserving utility in normal cases.
(3) Extensive experiments demonstrate that TAS achieves better safety-utility performance compared with eleven defense baselines on three benchmarks.

\section{Related Work}
\subsection{LLM Safety}
% Safety approaches for LLMs generally fall into two paradigms: post-training alignment and prompt engineering.
LLM safety approaches are roughly categorized into: Prompt Engineering and Post-training Alignment.
\textbf{(1) Prompt Engineering.}
Prompt engineering provides a lightweight defense by refining the input prompt to steer LLM behavior during inference, helping to avoid unsafe or undesired outputs. 
% Defense mechanisms often involve optimizing system prompts or decoding strategies.
\citet{xiong2025defensive} designed suffix prompts that adaptively minimize jailbreak success rates while maintaining the utility of LLMs.
However, such prompts often require careful design for each task, which limits their generalizability.
\citet{zhou2024robust} proposed an optimization-based algorithm for defending LLMs against jailbreak attacks, improving robustness against unseen jailbreaks.
\citet{li2025securitylingua} employed prompt compression to filter harmful instructions. 
Beyond input modification, \citet{xu2024safedecoding} modified the decoding probability distribution to identify safety tokens  and amplify their probabilities during inference. 
\citet{zhang2024defending} integrated goal prioritization during both training and inference to minimize the conflict between helpfulness and safety.
% proposed explicitly structure prompts to ensure safety while being helpful. 
% Despite their low resource cost, these methods are often static and they lack the temporal context required to secure the multi-turn workflows characteristic of autonomous agents.
However, they are vulnerable to adversarial prompt attacks, which diminish LLM safety by exploiting weaknesses in the input prompt.
\textbf{(2) Post-training Alignment.}
Post-training alignment is the set of techniques applied after pretraining to ensure LLMs refuse harmful queries.
% to internalize safety norms into model parameters.
% Standard alignments such as SFT and RLHF have been adapted to reduce harmful content generation. For example, 
\citet{chaudhary2024risk} introduced risk-averse RLHF to reduce harmful content generation during LLM fine-tuning and improve model safety. 
\citet{mu2024rule} proposed Rule-Based Rewards (RBR), which used rules for desired or undesired behaviors with an LLM grader, improving safety while reducing unnecessary refusals.
%%%%%%参考以上部分%%%%%%%%
% To address the data bottleneck, 
%\citet{ji-etal-2025-pku} propose PKU-SafeRLHF dataset, training models to better identify and reject unsafe content. 
% \citet{ji-etal-2025-pku} proposed PKU-SafeRLHF dataset, which separates annotations of helpfulness and harmlessness for question-answering pairs, providing a valuable resource on safety alignment in LLMs.
%Similarly, \citet{shi-etal-2024-safer} automates the construction of preference data to fine-tune models for harmlessness without massive manual annotation.
% 
% \citet{shi-etal-2024-safer} introduced Safer-Instruct to construct large-scale preference data for RLHF, which further improves LLM safety while maintaining a balanced level of task performance.
% 
% reduced limitations of existing automatic generation methods in both data diversity and quality.
%Recent work focuses on optimizing the alignment process itself.
% 
% Apart from improving the quality of training data, more recent works concentrated on advancing alignment techniques.
% 
% Advanced optimization frameworks, including Mask-DPO \citep{gu2025mask} and GC-DPO \citep{maeng-etal-2025-goal}, further refined this process by emphasizing factuality and prioritizing the system prompt.
%\citet{maeng-etal-2025-goal} enforce a priority hierarchy where system safety instructions override user prompts to defend against jailbreaks. 
%GC-DPO\citep{maeng-etal-2025-goal} reached a good balance between performance and safety by prioritizing the system prompt over the user prompt through goal-conditioning.
\citet{gu2025mask} introduced a Mask-DPO method, which incorporates factuality masking to eliminate noise in preference data, enhancing the universality of the alignment. 
% Furthermore, efficient alignment techniques such as
%\citet{gu2025mask} introduced Mask-DPO, which incorporated sentence-level factuality as mask signals to eliminate noise data in preference learning.
%For instance, Mask-DPO \citep{gu2025mask} incorporated sentence-level factuality as mask signals to eliminate noise in preference data and 
\citet{maeng-etal-2025-goal} proposed a GC-DPO method, which prioritizes the system prompt over the user prompt based on goals, resulting in a better balance between performance and LLM safety.
\citet{hsu2024safe} introduced a Safe LoRA method,  which projects updates into a safe subspace to alleviate safety degradation during fine-tuning.
Aligner \citep{NEURIPS2024_a51a74b2} predicted accurate residuals to repair unsafe outputs without adjusting model parameters. 

\subsection{Agent Safety}
In contrast to LLM single-step generation, agents operate through a multi-step workflow, requiring safety that spans the entire workflow.
The agent safety methods are divided into: Internal Alignment and External Verification.
\textbf{(1) Internal Alignment.}
Internal alignment approaches aim to give agents the ability to regulate their own behavior.
\citet{tennant2025moral} proposed reward functions for moral alignment of LLM agents, encouraging agent behaviors to better align with human values.
% which enable agents to provide a general solution to align LLM agents to human values.
\citet{NEURIPS2024_d0322637} introduced a World Knowledge Model (WKM), which provides global priors to facilitate agent planning and reduce hallucinatory actions, improving the agent's safety. 
% Checking as Context trains models to internalize a ``generate-critique-correct" loop, improving zero-shot safety.
However, relying solely on internal alignment is risky, since LLMs fail to self-correct reasoning errors without external feedback \citep{huang2024large}. 
%Furthermore, purely internal safeguards can be bypassed by taint-style vulnerabilities, so AgentFuzz is proposed for automatic detection of these threats \citep{liu2025make}.
\textbf{(2) External Verification.}
To further improve agent safety, recent work introduced external verification strategies.
\citet{xiang2024guardagent} deployed GuardAgent, a guardrail agent that protects target agents by verifying whether their behavior is safe.
\citet{shi2025progentprogrammableprivilegecontrol} proposed Progent, a programmable privilege control mechanism that enforces fine-grained tool-call policies to restrict agents to necessary actions.
\citet{li2026drift} introduced DRIFT, a dynamic rule-based isolation framework that combines secure planning, dynamic validation, and injection isolation to defend LLM agents against prompt injection attacks.
\citet{bhagwatkar2025indirect} studied firewall-based defenses for indirect prompt injections, using tool-input and tool-output firewalls at the agent--tool interface.
\citet{hu2025agentsentinelendtoendrealtimesecurity} proposed AgentSentinel, an end-to-end real-time security defense framework that intercepts sensitive operations and audits execution traces for computer-use agents.
% \citet{dou-etal-2025-multi} introduced an MPLSandbox method, which provides an isolated sub-sandbox to ensure agent-generated code safety. 
% Additionally, \citet{miculicich2025veriguard} proposed VeriGuard, which generates verifiable specifications alongside code.
Although these methods consider non-LLM components in the workflow, they typically treat each step in isolation or rely on local tool-level checks, and thus overlook the risk propagation inherent in multi-step processes.

\begin{figure*}[t]
    \centering
    \includegraphics[width=1.0\textwidth]{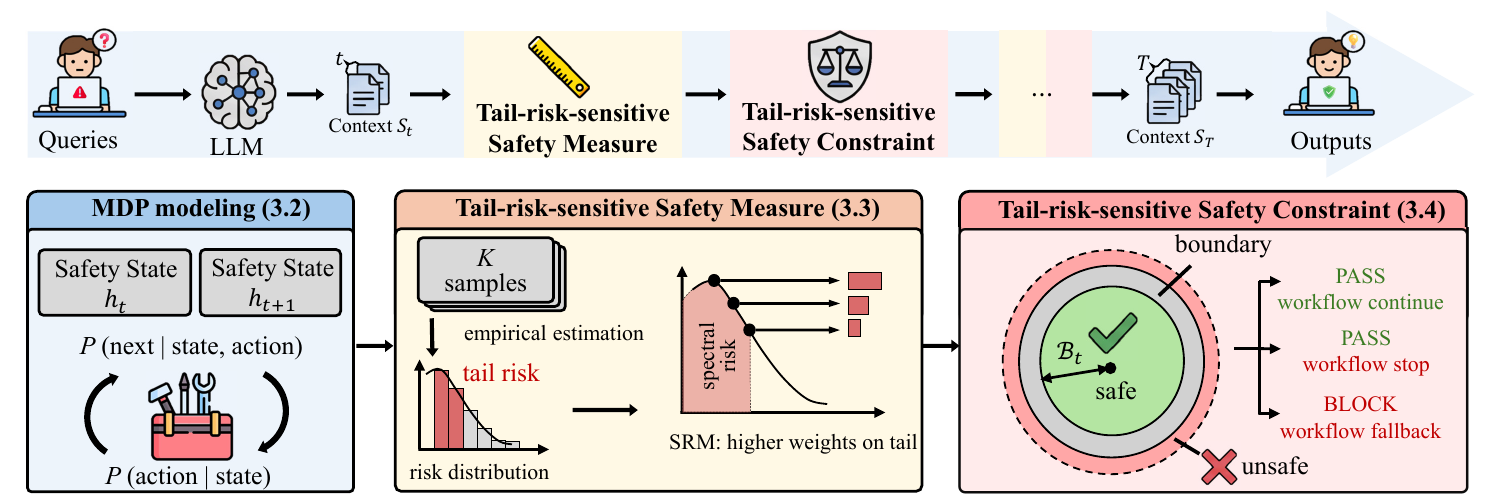}    
    \caption{Overview of TAS. The upper pipeline shows an LLM-based agent executing a multi-step workflow from queries to outputs. As shown in the bottom half, TAS models the safety state along the workflow as an MDP (blue left part), which formalizes the action-conditioned safety-state transitions. Then, with a candidate action, TAS samples candidate outcomes to estimate a risk distribution and measures the tail risk with SRM (yellow middle part). Next, TAS sets a shared budget constraint to check the risk score and provides a safety strategy under three cases (red right part). The process repeats until the workflow ends or the budget is exhausted, to finally generate outputs. 
    }
    \label{fig:main}
    % \vspace{-5pt}
\end{figure*}

\section{Methodology}
\label{3}
\subsection{Overview}
\tzlconfirmed{Our method consists of three parts:}
% (1) Modeling Agent Workflow as a Sequential Decision Process (\S~\ref{3.2}), which adopts the Markov Decision Process (MDP) to cast a sequential multi-step agent workflow as a sequential decision process, tracing risk transitions along the workflow;
% 
% (1) Modeling Agent Workflow as a Sequential Decision Process (\S~\ref{3.2}), which adopts a Markov Decision Process (MDP) to cast a sequential multi-step agent workflow, where the next step is conditioned on the current state, naturally as a sequential decision process, tracing how risk evolves across steps along the workflow;
% 
% (\S~\ref{3.2})
% (\S~\ref{3.3})
% (\S~\ref{3.3})
% (\S~\ref{3.4})
\tzlconfirmed{(1) Safety-State Modeling as a Sequential Decision Process (\S~\ref{3.2}), which introduces an agent safety state for each workflow step and formulates safety-state transitions along the workflow as a Markov Decision Process (MDP), so that each decision is conditioned on the current safety state and we can track how risk propagates across the workflow;}
% 
% (2) Safety Tail Boundaries via Spectral Risk Constraints (\S~\ref{3.3}), \tzlconfirmed{which introduces a safety constraint for risk} transitions via Spectral Risk Measure (SRM) to \tzlconfirmed{set a safety boundary explicitly for those rare but high-risk cases (i.e., tail); and}
% 
\tzlconfirmed{(2) Tail-risk-sensitive Safety Measures via Spectral Risk (\S~\ref{3.3}), which introduces a safety measure that targets rare but high-risk cases (i.e., tail) by assigning larger weights to high-risk outcomes than to common low-risk ones;}
% 
% (3) \tzlconfirmed{Tail-Constrained Optimization via Lagrangian Relaxation (\S~\ref{3.4}), which} provides the solutions for tail-constrained MDP with a Lagrangian formulation and yields a corresponding safety strategy for the current workflow step.
% 
% \tzlconfirmed{(3) Tail-constrained Safety Strategy with Empirical Estimate (\S~\ref{3.4}), which optimizes the tail-constrained MDP by relaxing the spectral risk constraint into a Lagrangian objective and empirically estimates a risk distribution to decide a safety strategy for the current workflow step.
\tzlconfirmed{(3) Tail-risk-sensitive Safety Constraint with Empirical Estimation (\S~\ref{3.4}), which introduces a shared budget constraint for tail risk by empirically estimating from sampling and applies a step-wise safety strategy.}
% \tzlconfirmed{Overall, TSA first formalizes the sequential agent safety state along the workflow as an MDP to trace risk propagation, then sets a tail-sensitive constraint for tail risks, and finally optimizes the workflow under the constraint to provide a safety strategy for the current step.}
\tzlconfirmed{Overall, TAS first formalizes the sequential agent safety state along the workflow (\S~\ref{3.2}) to trace risk propagation, then measures a tail-sensitive risk (\S~\ref{3.3}), and finally constrains the risk with a shared budget to derive a step-wise safety strategy (\S~\ref{3.4}).
% finally derives a step-wise safety strategy for the current step under a shared budget constraint (\S~\ref{3.4}).
}

% \vspace{-2mm}
% \subsection{Modeling Agent Workflow as a Sequential Decision Process}
\subsection{Safety-State Modeling as A Sequential Decision Process}
\label{3.2}
\tzlconfirmed{To formally model the sequential safety state in an agent workflow, we adopt the MDP, which regards each step in the workflow as a controlled safety state transition, depending on both the current safety state and the action choice, to trace risk propagation under different sequential decisions.}
% which serves as a mathematical framework \tzlconfirmed{for the sequential process.} 
% 
% \tzlconfirmed{The motivation is that existing methods for agent safety typically ignore internal dependencies across steps in the workflow, resulting in an unsafe agent with risk propagation.
% Since the agent workflow is inherently sequential, we argue that risk propagates depending on action decisions and outputs from the last step \cite{yao2023react,xi2023risepotentiallargelanguage}.}
% 
\tzlconfirmed{
% The motivation is that MDP adds an explicit action decision as conditions for the next step ($P(h_{t+1}\mid h_t,a_{t+1})$ rather than a normal sequential transition that predicts the next step from the current output alone (i.e., $P(s_{t+1}\mid s_t)$).
The motivation is that MDP adds an explicit action decision as a condition for the next safety-state transition, i.e., $P(h_{t+1}\mid h_t,a_{t+1})$, rather than a normal sequential transition that predicts the next state from the current state alone, i.e., $P(h_{t+1}\mid h_t)$.
Since the agent workflow is inherently sequential with multi-step actions, we argue that risk propagates depending on action decisions and accumulated workflow observations \cite{yao2023react,xi2023risepotentiallargelanguage}.
% and outputs from the last step \cite{yao2023react,xi2023risepotentiallargelanguage}.
Therefore, applying an MDP to model the action-conditioned risk propagation makes sense.}

\tzlconfirmed{Specifically, let $\mathcal{H}$ denote the space of agent safety states, and let $h_t \in \mathcal{H}$ be the observable safety state at step $t$, including initial user query $x$, accumulated workflow observations $s_t$ from agent possible actions (i.e., memory, retrievals, planning, LLM generations, tool-call results, and observable environment feedback), and accumulated safety cost 
% $C_t=\sum_{k=0}^{t-1}c_t$
$C_t$
: $h_t \triangleq (x, s_t, C_t)$.}
% 
% At each step, given the current safety state $h_t$, the agent first produces (1) an action $a_{t+1}$ from the action space $\mathcal{A}$: $a_{t+1} \in \mathcal{A}$ to select a module or tool to execute with $P_\theta(a_{t+1} \mid h_t)$, and then produces (2) a step-wise output $\tilde{s}_{t+1}$ generated under the chosen action $a_{t+1}$ with $P_\theta(\tilde{s}_{t+1} \mid h_t, a_{t+1})$.
% 
By aggregating these observable workflow signals, the safety state provides the information TAS uses to estimate how risk evolves across subsequent actions.
\tzlconfirmed{At each step, the agent conditions on the current safety state $h_t$ and makes a decision in two stages.
As shown in Eq.~\eqref{eq:joint_distribution}, it first chooses an action $a_{t+1}\in\mathcal{A}$ (e.g., which module/tool to execute) according to $P_\theta(a_{t+1}\mid h_t)$. 
Second, given the chosen action, it generates a step-wise output $\tilde{s}_{t+1}$ according to $P_\theta(\tilde{s}_{t+1}\mid h_t,a_{t+1})$.
We model these two stages by their joint distribution as:
\begin{equation}
% \small
P_\theta(\tilde{s}_{t+1},a_{t+1}\mid h_t)
=
\underbrace{P_\theta(a_{t+1} \mid h_t)}_{\text{action decision}}
\cdot
\underbrace{P_\theta(\tilde{s}_{t+1} \mid h_t, a_{t+1})}_{\text{intermediate output}}.
\label{eq:joint_distribution}
\end{equation}
}
\tzlconfirmed{We then update $s_{t+1}=\mathrm{concat}(s_{t}, \tilde{s}_{t+1})$ and $C_{t+1}=C_t+c(h_t,a_{t+1},\tilde{s}_{t+1})$ to get the next safety state $h_{t+1}=(x,s_{t+1},C_{t+1})$.}

\tzlconfirmed{Based on the definition above, an agent workflow over $T$ steps induces a safety-state trace: $\tau \triangleq (h_0, a_1,\tilde{s}_1, h_1, \ldots, a_T, \tilde{s}_T, h_T)$
, starting from initial state $h_0=(x,s_0,C_0)$ and updating by action $a_t$ sequentially.
As shown in Eq.~\eqref{eq:trace}, given safety state $h_t$, the agent selects the action $a_{t+1}\sim P_\theta(a_{t+1}\mid h_t)$ and generates a step-wise intermediate output $\tilde{s}_{t+1}\sim P_\theta(\tilde{s}_{t+1}\mid h_t,a_{t+1})$.
As a result, the trace distribution follows the chain rule:
% \begin{equation}
% \small
% \label{eq:trace}
% \begin{aligned}
% P_{\pi_\theta}(\tau)
% &=
% P(h_0)\prod_{t=0}^{T-1} P_\theta(a_{t+1}, \tilde{s}_{t+1}\mid h_t)\\
% &=
% P(h_0)\prod_{t=0}^{T-1} P_\theta(a_{t+1}\mid h_t)\,P_\theta(\tilde{s}_{t+1}\mid h_t,a_{t+1}).
% \end{aligned}
% \end{equation}
\begin{equation}
\begin{aligned}
\small
\label{eq:trace}
P_{\pi_\theta}(\tau)
&=
P(h_0)\prod_{t=0}^{T-1} P_\theta(a_{t+1}, \tilde{s}_{t+1}\mid h_t)\\
&=
P(h_0)\prod_{t=0}^{T-1} P_\theta(a_{t+1}\mid h_t)\,P_\theta(\tilde{s}_{t+1}\mid h_t,a_{t+1}).
\end{aligned}
\end{equation}
Meanwhile, the accumulated safety cost is updated as $C_{t+1} = C_t + c(h_t,a_{t+1},\tilde{s}_{t+1})$ with the step-wise safety cost $c$. 
% defined in Eq.~\eqref{eq:step_cost}. 
Thus, we can get the next safety state $h_{t+1}=(x,s_{t+1},C_{t+1})$. 
We then trace risk propagation across steps in the workflow with MDP modeling for better safety measures in the following section.}
% 
% The MDP model adds the action decisions in the workflow as conditions for safety-state transitions, rather than a normal sequential model that predicts the next step only conditioned on the current output. Therefore, MDP can dependently trace action-conditioned risk propagation across steps for better safety measures in the following section.
% 
% A normal sequential model typically treats the workflow as a sequence of outputs, making it hard to model how different action decisions would dependently impact risk propagation.
% In contrast, our MDP modeling considers the action decisions in the workflow as conditions for safety-state transitions, allowing us to dependently trace risk propagation across steps.

% As shown in Eq.~\eqref{eq:trace}, malicious attacks can affect both action selection $P_\theta(a_{t+1}\!\mid h_t)$ and intermediate output generation $P_\theta(\tilde{s}_{t+1}\!\mid h_t,a_{t+1})$, producing unsafe action decisions or contents that propagate to next agent state $h_{t+1}$ (i.e., risk transition).
% Thus, LLM-only generation safety (focusing $\small P_\theta(\tilde{s}_{t+1}\!\mid h_t,a_{t+1})$) or isolated per-step checks (focusing $\small P_\theta(a_{t+1}\!\mid h_t)$) are insufficient for the entire agent workflow safety, as we mentioned previously.
% As a result, it motivates constraining the overall risk propagation with a single shared budget along the workflow as we present next.

\subsection{Tail-risk-sensitive Safety Measures via Spectral Risk}
\label{3.3}
% \tzlconfirmed{To prevent rare but high-risk failures in an agent workflow, we impose a tail-sensitive safety constraint with a single shared budget over dependent steps along the workflow instead of independent constraints for each step.
% The motivation is that measuring safety in an overall-average way can hide those rare but extreme cases, because most cases 
% % (e.g., 95\%) 
% are typically safe with near-zero cost, while a small fraction 
% % (e.g., 5\%) 
% incurs extreme cost (i.e., tail risk) \cite{bastani2022regretboundsrisksensitivereinforcement,lam2023riskaware}.
% Therefore, we argue that applying a tail-sensitive constraint to focus more on tail risks, which refers to low-probability but high-cost outcomes, can ensure a safer agent workflow.}

\tzlconfirmed{To assess rare but high-risk failures in an agent workflow, we apply a tail-sensitive safety measure via spectral risk measures (SRM) to target the tail risk instead of the overall average.
The motivation is that measuring safety in an overall-average way can hide those rare but extreme cases, because most cases are typically safe with near-zero cost, while a small fraction incurs extreme cost (i.e., tail risk) \cite{bastani2022regretboundsrisksensitivereinforcement,lam2023riskaware}.
Therefore, we argue that applying a tail-sensitive constraint to focus more on tail risks, which refers to low-probability but high-cost outcomes, can ensure a safer agent workflow.}

% Specifically, we note that expectation-based safety measures $\mathbb{E}_{\tau\sim P_{\pi_\theta}}[C(\tau)]$, where $C(\tau)$ is the cumulative cost along the safety-state trace in the workflow, can mask tail risk.
% 
\tzlconfirmed{Specifically, we note that expectation-based safety measures, like $\mathbb{E}_{\tau \sim P_{\pi_\theta}}[C(\tau)]$, may ignore tail risks because it focuses on the overall cumulative cost $C(\tau)$ along the safety-state trace.}
\tzlconfirmed{In practice, a small fraction of outcomes with low probability ($\epsilon$) may cost extremely high, while most of the steps with high probability ($1-\epsilon$) are benign and cost a little.
Due to the imbalanced long-tail risk distribution, the expectation-based safety measures $\mathbb{E}[C(\tau)] \approx \epsilon \cdot C_{\text{tail}}$ can still be small even if $C_{\text{tail}}\gg0$, which makes the entire workflow unsafe.}

\tzlconfirmed{
\noindent
\textbf{Step-wise safety cost.}
% To evaluate safety along the workflow, we assign a step-wise safety cost at each step $t$ as
% $c_t \triangleq c(h_t,a_{t+1},\tilde{s}_{t+1})$.
% We compute $c_t$ considering both the rule-based feature patterns (e.g., email, address, phone number, ID card, ...) and API-based semantic risks, with the detailed reproducible calculation provided in App.~\ref{app:step_cost}.
% We compute $c_t$ by weighted ($\gamma_{1,2}$) combining a normalized rule-based feature matching score $s^{\text{rule}}_t\in[0,1]$ (e.g., email, address, phone number, ID card, ...), and an API-based semantic risk score $s^{\text{api}}_t\in[0,1]$ (calculation details in App.~\ref{app:step_cost}):
% \begin{equation}
% c_t
% =
% \gamma_1\, s^{\text{rule}}_t + \gamma_2\, s^{\text{api}}_t.
% \label{eq:step_cost}
% \end{equation}
% where $s^{\text{rule}}_t\in[0,1]$ is a score produced by rule matching (e.g., email, address, phone number, ID card, ...),
% $s^{\text{api}}_t\in[0,1]$ is the risk score returned by an external safety API for the step output,
% and $\gamma_{1,2}$ denote their weights.
% 
To evaluate safety along the workflow, TAS computes a step-wise safety cost
\(c_t \triangleq c(h_t,a_{t+1},\tilde{s}_{t+1})\). The cost is based on two signals:
(i) a rule-based score that first pre-classifies the tool invocation from tool names,
arguments, PII/secret patterns, and external-target checks, and then computes a
nonlinear feature-based cost; and (ii) an API-based semantic scorer that evaluates
the transition from the tool name, truncated arguments, extracted rule-based features,
and current workflow context, returning a semantic risk score. This semantic score is combined with the rule-based cost to produce a budget-scale final \(c_t\). Full scoring rules, prompts, and implementation details are provided in App.~\ref{app:step_cost}.
We summarize the step-wise cost to get the cumulative workflow cost as:
\begin{equation}
C(\tau) \triangleq \sum_{t=0}^{T-1} c_t, \qquad \tau\sim P_{\pi_\theta}.
\label{eq:workflow_cost}
\end{equation}
}
\vspace{-4mm}

\tzlconfirmed{
\noindent
\textbf{Tail-sensitive risk via SRM.}
To precisely quantify tail risk severity, we employ Spectral Risk Measures (SRM) \cite{NEURIPS2024_a7fbf054,moghimi2025cvarleveragingstaticspectral} with a monotonically increasing weighting function, assigning higher weights to higher risk.
Based on the cumulative safety cost $C(\tau)$ of a workflow trace $\tau$ under $\tau \sim P_{\pi_\theta}$, we quantify tail risk by taking a weighted average over the $u$-quantiles of $C(\tau)$.
We denote the $u$-quantile by $F_C^{-1}(u), u\in[\alpha,1]$, in which $u$ is the quantile, $\alpha\in(0,1)$ is a quantile cutoff for tail region boundary, and $F_C$ is the cumulative distribution function of $C(\tau)$.
We assign a nonnegative and monotonically increasing weight $\sigma(u)$, which increases with $u$, so that higher-cost outcomes receive larger weights.
}
% 
% normalized with $\int_0^1 \sigma(u)\,du=1$
\tzlconfirmed{We normalize $\sigma$ with $\int_\alpha^1 \sigma(u)\,du=1$ so that $M_\sigma$ remains on the same scale as $C(\tau)$ and make it comparable to the budget $\mathcal{B}$.}
\tzlconfirmed{As a result, we can obtain a tail-sensitive risk score $M_\sigma(\cdot)$ by integrating these weighted $u$-quantiles of cumulative cost $C(\tau)$:
% \vspace{-2mm}
\begin{equation}
\label{eq:risk_score}
    M_\sigma\!\big(C(\tau)\big) \triangleq \int_\alpha^1 \sigma(u)\,F_C^{-1}(u)\,du.
\end{equation}
}
\vspace{-4mm}

\tzlconfirmed{Above all, we explicitly measure the tail risk to focus on those rare but high-risk failures, which are typically hidden in an overall average.
Therefore, we focus on constraining the tail risk for a step-wise safety strategy in the following.}

\begin{table*}[t]
\centering
\footnotesize
\setlength{\tabcolsep}{2.7pt}
\renewcommand{\arraystretch}{0.92}

% Compact spacing around booktabs rules, local to this table.
\begingroup
\setlength{\aboverulesep}{1.1pt}
\setlength{\belowrulesep}{1.1pt}
\setlength{\cmidrulesep}{1.0pt}

\begin{tabular*}{\textwidth}{
    @{\extracolsep{\fill}}
    l l
    *{11}{c}
    @{}
}
\toprule
\multirow{2}{*}{\textbf{Model}}
&
\multirow{2}{*}{\textbf{Metric}}
&
\multicolumn{6}{c}{\textbf{DPI}}
&
\multicolumn{5}{c}{\textbf{IPI}}
\\
\cmidrule(lr){3-8}
\cmidrule(lr){9-13}
&
&
\textbf{No Def.}
&
\textbf{Deli.}
&
\textbf{Para.}
&
\textbf{Instr.}
&
\textbf{Rewr.}
&
\textbf{Ours}
&
\textbf{No Def.}
&
\textbf{Deli.}
&
\textbf{Instr.}
&
\textbf{Sand.}
&
\textbf{Ours}
\\
\midrule

\multirow{3}{*}{Llama3-70B}
& ASR $\downarrow$
& 87.75
& 88.25
& \underline{71.25}
& 87.25
& 74.50
& \textbf{43.50}
& 45.50
& \underline{34.50}
& 41.50
& 39.75
& \textbf{13.95}
\\
& NRP $\uparrow$
& 6.80
& 6.76
& \underline{14.81}
& 7.87
& 12.69
& \textbf{32.91}
& 30.25
& \underline{37.66}
& 36.12
& 32.54
& \textbf{50.12}
\\
& PNA $\uparrow$
& 55.50
& 57.50
& 51.50
& 61.75
& 49.75
& 58.25
& 55.50
& 57.50
& 61.75
& 54.00
& 58.25
\\
\midrule

\multirow{3}{*}{Qwen2-72B}
& ASR $\downarrow$
& 94.50
& 95.00
& 60.50
& 95.50
& \underline{56.20}
& \textbf{48.35}
& 23.75
& \underline{17.50}
& 26.50
& 21.75
& \textbf{17.05}
\\
& NRP $\uparrow$
& 1.09
& 1.06
& \underline{8.69}
& 1.05
& 7.77
& \textbf{11.36}
& 15.06
& \underline{17.53}
& 17.09
& 16.82
& \textbf{18.25}
\\
& PNA $\uparrow$
& 19.75
& 21.25
& 22.00
& 23.25
& 17.75
& 22.00
& 19.75
& 21.25
& 23.25
& 21.50
& 22.00
\\
\midrule

\multirow{3}{*}{GPT-4o}
& ASR $\downarrow$
& 55.50
& \underline{52.25}
& 62.50
& 70.75
& 60.20
& \textbf{40.35}
& 62.00
& 66.75
& \underline{61.75}
& 64.75
& \textbf{45.65}
\\
& NRP $\uparrow$
& \underline{35.60}
& 33.31
& 29.91
& 20.11
& 31.34
& \textbf{47.42}
& \underline{30.40}
& 23.19
& 26.30
& 28.02
& \textbf{43.21}
\\
& PNA $\uparrow$
& 80.00
& 69.75
& 79.75
& 68.75
& 78.75
& 79.50
& 80.00
& 69.75
& 68.75
& 79.50
& 79.50
\\
\midrule

\multirow{3}{*}{GPT-5.2}
& ASR $\downarrow$
& 69.80
& 64.45
& 61.40
& \underline{53.85}
& 69.25
& \textbf{48.40}
& 63.15
& 65.35
& \underline{55.00}
& 58.30
& \textbf{42.25}
\\
& NRP $\uparrow$
& 27.78
& 33.51
& 36.57
& \underline{42.34}
& 28.52
& \textbf{48.89}
& 33.90
& 32.66
& \underline{41.29}
& 39.09
& \textbf{54.72}
\\
& PNA $\uparrow$
& 92.00
& 94.25
& 94.75
& 91.75
& 92.75
& 94.75
& 92.00
& 94.25
& 91.75
& 93.75
& 94.75
\\
\midrule

\multirow{3}{*}{\textbf{Avg.}}
& ASR $\downarrow$
& 76.89
& 74.99
& \underline{63.91}
& 76.84
& 65.04
& \textbf{45.15}
& 48.60
& \underline{46.03}
& 46.19
& 46.14
& \textbf{29.73}
\\
& NRP $\uparrow$
& 14.29
& 15.18
& \underline{22.37}
& 14.22
& 20.89
& \textbf{34.90}
& 31.77
& 32.76
& 33.03
& \underline{33.50}
& \textbf{44.71}
\\
& PNA $\uparrow$
& 61.81
& 60.69
& 62.00
& 61.38
& 59.75
& 63.63
& 61.81
& 60.69
& 61.38
& 62.19
& 63.63
\\
\bottomrule
\end{tabular*}

\endgroup

\caption{
Main results of five defenses on four LLM backbones under DPI and IPI attacks. Lower ASR indicates better safety, while higher PNA and NRP indicate better utility and overall performance. We focus on ASR and NRP, with PNA reported to ensure utility remains close to \emph{No Defense}. Best and second-best results are bolded and underlined. All improvements are significant under a two-sided exact McNemar test ($p<0.01$) (App.~\ref{app:sig}).
}
\label{tab:main_results}
\end{table*}

\begin{table}[t]
\centering
\small
\setlength{\tabcolsep}{3pt}
\renewcommand{\arraystretch}{0.94}

\begingroup
\setlength{\aboverulesep}{0.9pt}
\setlength{\belowrulesep}{0.9pt}

\begin{tabular*}{\columnwidth}{
    @{\extracolsep{\fill}}
    lcccc
    @{}
}
\toprule
\textbf{Model}
&
\textbf{No Def.}
&
\textbf{Shuffle}
&
\textbf{Para.}
&
\textbf{Ours}
\\
\midrule

Llama3-70B
& 57.00
& 63.75
& \underline{44.75}
& \textbf{15.00}
\\

Qwen2-72B
& 96.50
& 81.50
& \underline{57.75}
& \textbf{43.25}
\\

GPT-4o
& 100.00
& 98.50
& \underline{84.75}
& \textbf{42.75}
\\

GPT-5.2
& 87.25
& 88.50
& \underline{67.00}
& \textbf{39.75}
\\

\midrule

\textbf{Avg.}
& 85.19
& 83.06
& \underline{63.56}
& \textbf{35.19}
\\

\bottomrule
\end{tabular*}
\endgroup

\caption{ASR (\%) $\downarrow$ under POT attacks across four LLM backbones.}
\label{tab:main_pot}
\end{table}

\begin{table*}[t]
\centering
\footnotesize
\setlength{\tabcolsep}{2.6pt}
\renewcommand{\arraystretch}{1.00}

\begin{tabular}{@{}lllccccccc@{}}
\toprule
\textbf{Benchmark}
&
\textbf{Metric}
&
\textbf{Model}
&
\textbf{No Def.}
&
\textbf{Progent}
&
\textbf{DRIFT}
&
\textbf{Firewall}
&
\textbf{Sentinel}
&
\textbf{GuardAgent}
&
\textbf{Ours}
\\
\midrule

\multirow{4}{*}{\shortstack{Agent-\\SafetyBench}}
&
\multirow{4}{*}{\shortstack{Safety \\Score $\uparrow$}}
&
Llama3-70B
&
0.325
&
0.400
&
0.383
&
0.408
&
0.492
&
0.325
&
\cellcolor{oursgray}\textbf{0.508}
\\

&
&
Qwen2-72B
&
0.225
&
0.275
&
0.292
&
0.275
&
0.383
&
\textbf{0.483}
&
\cellcolor{oursgray}\textbf{0.483}
\\

&
&
GPT-4o
&
0.442
&
0.442
&
0.450
&
0.417
&
0.700
&
0.450
&
\cellcolor{oursgray}\textbf{0.758}
\\

&
&
GPT-5.2
&
0.625
&
0.633
&
0.667
&
0.592
&
0.808
&
0.675
&
\cellcolor{oursgray}\textbf{0.908}
\\

\midrule

\multirow{4}{*}{AgentDojo}
&
\multirow{4}{*}{\shortstack{ASR $\downarrow$\\{/} Utility $\uparrow$}}
&
Llama3-70B
&
0.18 / 10.98
&
0.36 / 10.90
&
0.00 / 10.37
&
0.00 / 11.11
&
0.00 / 9.53
&
0.00 / 11.03
&
\cellcolor{oursgray}\textbf{0.00 / 11.11}
\\

&
&
Qwen2-72B
&
0.00 / 9.15
&
0.00 / 8.83
&
0.00 / 8.54
&
0.00 / 9.15
&
0.00 / 7.35
&
0.00 / 9.15
&
\cellcolor{oursgray}\textbf{0.00 / 9.15}
\\

&
&
GPT-4o
&
47.51 / 56.74
&
46.51 / 62.72
&
15.81 / 30.60
&
12.74 / 52.79
&
4.96 / 10.19
&
5.04 / 34.93
&
\cellcolor{oursgray}\textbf{3.93 / 63.32}
\\

&
&
GPT-5.2
&
6.78 / 76.91
&
6.13 / \textbf{76.92}
&
5.09 / 41.53
&
4.86 / 64.34
&
4.41 / 14.59
&
1.41 / 50.04
&
\cellcolor{oursgray}\textbf{1.36} / 76.43
\\

\bottomrule
\end{tabular}

\caption{
Additional results on Agent-SafetyBench and AgentDojo across four
LLM backbones. %AgentDojo cells report ASR / Utility.
}
\label{tab:combined-benchmark-results}
\end{table*}
%%%%%%%%%%%%%%%%%%%%%%%%%%%%%%%%%%%%%%%%

% \subsection{Tail-Constrained Optimization via Lagrangian Relaxation}
\subsection{Tail-risk-sensitive Safety Constraint with Empirical Estimation}
\label{3.4}
\tzlconfirmed{
% To prevent tail risk from propagating along the agent workflow, we set a shared safety budget to constrain the risk, which is empirically estimated with sampling.
% 
To control tail-risk propagation along the agent workflow, we maintain a shared safety budget and update it online as the workflow proceeds.
The motivation is that an individual constraint for a single step cannot trace the accumulated risk in the entire workflow \cite{xiang2024guardagent,huang2024large}, so we share a safety budget across steps to prevent risk propagation.
Additionally, since the practical risk distribution for the tail-risk measure is unknown, we empirically approximate it with sampling.}

\tzlconfirmed{
\noindent
\textbf{Shared workflow-level budget.}
We ensure agent workflow safety by constraining the tail-sensitive risk score $M_\sigma$ within a single shared budget $\mathcal{B}$: $M_\sigma\!\big(C(\tau)\big) \le \mathcal{B}$ (Eq.~\eqref{eq:risk_score}).
% \begin{equation}
% M_\sigma\!\big(C(\tau)\big) \le \mathcal{B}.
% \label{eq:budget}
% \end{equation}
In practice, we extend the static budget check $\mathcal{B}$ as a dynamic online budget with the workflow proceeding.
We track the accumulated safety cost $C_t$ in the safety state and focus on the remaining budget $B_t \triangleq \mathcal{B}-C_t$. 
}
\tzlconfirmed{Therefore, we define the incremental cost $\Delta C_{t+1}\triangleq c(h_t,a_{t+1},\tilde{s}_{t+1})$, where the step-wise output $\tilde{s}_{t+1}$ is generated conditioned on safety state and action $(h_t,a_{t+1})$ as in \S\ref{3.2}.
We then check whether the tail risk fits within the remaining budget $B_t$:}
\begin{equation}
\small
M_\sigma\!\big(\Delta C_{t+1}\mid h_t,a_{t+1}\big)
~\triangleq~
\int_{\alpha}^{1}\sigma(u)\,F^{-1}_{\Delta C\mid h_t,a_{t+1}}(u)\,du
\ \le\ B_t,
\label{eq:dynamic_budget}
\end{equation}
\tzlconfirmed{
% In this way, high-risk steps that would exhaust the remaining tail-risk budget are rejected from propagating risks to later steps.
% Unlike an individual threshold for a single step, $\mathcal{B}$ controls the overall tail risk of the entire workflow distribution under $P_{\pi_\theta}$, thus accounting for inter-step risk propagation.
Unlike independent thresholds for individual steps, $\mathcal{B_t}$ is updated throughout the workflow and couples the safety decisions across steps. Earlier risky transitions reduce the remaining budget for later steps, allowing TAS to control risk propagation across the workflow.
}

\noindent
\textbf{Empirical estimation of risk distribution.}
\tzlconfirmed{To approximate the unknown practical distribution of safety costs (i.e., $F_{\Delta C\mid h_t,a_{t+1}}$ in Eq.~\eqref{eq:dynamic_budget}), we empirically estimate the distribution with sampling.
As shown in Eq.~\eqref{eq:mc_samples}, we repeatedly draw $K$ independent candidate outputs $\{\tilde{s}_{t+1}^{(k)}\}_{k=1}^K$ under the same state and action $(h_t,a_{t+1})$:}
\begin{equation}
\tilde{s}_{t+1}^{(k)} \sim P_\theta(\cdot \mid h_t,a_{t+1}), \quad k=1,\ldots,K, 
\label{eq:mc_samples}
\end{equation}
% with 0.7 temperature and top-$p=0.95$
\tzlconfirmed{and compute the corresponding costs: $c_t^{(k)} \triangleq c\!\left(h_t,a_{t+1},\tilde{s}_{t+1}^{(k)}\right).$
We sort the sample costs in nondecreasing order $c_{t,(1)} \le \cdots \le c_{t,(K)}$.
We then apply the SRM in Eq.~\eqref{eq:risk_score} 
% with quantile cutoff $\alpha$ 
to compute the empirical tail-risk costs, which focuses on the tail samples after $\lceil \alpha K\rceil$
% (i.e., sorted index after $\lceil \alpha K\rceil$) 
with higher costs:}
% \begin{equation}
% \small
% \widehat{M}_{\sigma,t}
% \triangleq
% \sum_{k=\lceil \alpha K\rceil}^K w_k\,c_{t,(k)},
% \quad \sum_{k=\lceil \alpha K\rceil}^{K} w_k = 1,\ w_k\ge 0,
% \label{eq:emp_cost}
% \end{equation}
\begin{equation}
\small
\widehat{M}_{\sigma,t}
\triangleq
\sum_{k=\lceil \alpha K\rceil}^K w_k\,c_{t,(k)},
\quad
w_k =
\frac{\sigma\!\left(\frac{k}{K}\right)}
{\sum_{j=\lceil \alpha K\rceil}^{K} \sigma\!\left(\frac{j}{K}\right)}.
\label{eq:emp_cost}
\end{equation}
% \tzlconfirmed{where $\{w_k\}_{k=\lceil \alpha K\rceil}^K$ are the normalizing weights from $\sigma(\cdot)$ in Eq.~\eqref{eq:dynamic_budget}.}
where $k=\lceil \alpha K\rceil,\ldots,K $ and $w_k$ is the normalized weights induced by the non-negative spectral function $\sigma(\cdot)$.
% where the discrete weights are obtained by normalizing the spectral weights on the sampled quantile grid:
% $
% w_k =
% \frac{\sigma\!\left(\frac{k}{K}\right)}
% {\sum_{j=\lceil \alpha K\rceil}^{K} \sigma\!\left(\frac{j}{K}\right)},
% \quad k=\lceil \alpha K\rceil,\ldots,K .
% $

\noindent
\textbf{Tail-constrained safety strategy.}
\tzlconfirmed{Based on the budget constraint in Eq.~\eqref{eq:dynamic_budget} and the empirical tail-risk estimation in Eq.~\eqref{eq:emp_cost}, TAS aims to preserve task utility while keeping the tail risk within the constraint.
% our goal is to learn a policy that achieves high utility while keeping the tail risk within the constraint. 
Let $\tau$ denote the safety-state trace generated by the policy $\pi$ as $\tau \sim P_\pi$, and let $C(\tau)$ be its cumulative safety cost as Eq.~\eqref{eq:workflow_cost}.
We define the utility objective as $J_{\text{util}}(\pi) \triangleq \mathbb{E}_{\tau\sim P_\pi}[U(\tau)]$, where $U(\tau)$ denotes the task utility accumulated along the trace (e.g., original task success rate).
Therefore, 
% as shown in Eq.~\eqref{eq:primal_cmdp}, 
our strategy goal is to maximize the $J_{\text{util}}(\pi)$ within the constraint $\widehat{M}_{\sigma,t} \le \mathcal{B}_t$ as:}
% \begin{equation}
% \small
% \label{eq:primal_cmdp}
$
\max_{\pi} \; J_{\text{util}}(\pi)
\quad
\text{s.t.}\quad
\widehat{M}_{\sigma,t} - \mathcal{B}_t\le0,\ \ \forall t.
$
% \end{equation}
\tzlconfirmed{Therefore, the budget check yields three practical cases for the current step:
\textbf{(1) Budget satisfied.} If $\widehat{M}_{\sigma,t}<\mathcal{B}_t$, the estimated tail risk is below the remaining budget. In this case, we allow the step to prioritize utility.
\textbf{(2) On the budget boundary.} If $\widehat{M}_{\sigma,t}=\mathcal{B}_t$, the step exactly uses up the remaining budget. We allow this step, but stop the workflow afterward to avoid any further risk.
\textbf{(3) Budget violated.} 
% If $\widehat{M}_{\sigma,t}>\mathcal{B}_t$, executing this step would exceed the budget. We reject and rollback the step to prevent risk propagation.
If $\widehat{M}_{\sigma,t} > B_t$, executing this step would exceed the budget. TAS rejects the step, rolls back to the pre-action safety state $h_t$, and masks the rejected action. It then resamples an alternative action and applies the same budget check.
% We limit this process to at most $R$ retries. 
If all retries (e.g., 3 times) fail, TAS returns a deterministic safe fallback response and terminates the workflow.
}

\section{Experiments}

\subsection{Experimental Settings}
\label{4.1}
\noindent\textbf{Models\footnote{We change LLM backbones in the LLM-based agent workflows, while keeping other parts the same.}.}
\tzlconfirmed{We use four open-source LLMs, including LlaMA3-8B-Instruct, LlaMA3-70B-Instruct, Qwen2-7B-Instruct, and Qwen2-72B-Instruct (additional results of the small LLMs in App.~\ref{app:exp_full_main}). We also use two closed-source LLMs, including GPT-4o and GPT-5.2.}

 % App.~\ref{app:exp_full_main}

% \noindent\textbf{Datasets.} 
% \tzlconfirmed{We use Agent Security Bench (ASB) \cite{zhang2025agent} to evaluate our method and all other baselines.}
\noindent\textbf{Benchmarks.}
We use three benchmarks for evaluations, including Agent Security Bench (ASB) \cite{zhang2025agent}, Agent-SafetyBench \cite{zhang2024agent}, and AgentDojo \cite{debenedetti2024agentdojo}.
\noindent\textbf{Baselines.}
We compare TAS with baselines under the corresponding attack settings in ASB. For DPI and IPI attacks, we include prompt-injection defenses, including \texttt{Deli.}~\cite{learning_prompt_data_isolation_url, mattern2023membership, pi_against_gpt3}, \texttt{Para.}~\cite{jain2023baseline}, \texttt{Instr.}~\cite{learning_prompt_instruction_url}, \texttt{Rewr.}~\cite{zhang2025agent}, and \texttt{Sand.}~\cite{learning_prompt_sandwich_url}. For POT backdoor attacks, we compare with \texttt{Para.} and \texttt{Shuffle}~\cite{xiang2023cbd}. We also compare with recent agent-safety defenses, including \texttt{Progent}~\cite{shi2025progentprogrammableprivilegecontrol}, \texttt{DRIFT}~\cite{li2026drift}, \texttt{Firewall}~\cite{bhagwatkar2025indirect}, \texttt{GuardAgent}~\cite{xiang2024guardagent}, and \texttt{Sentinel}~\cite{hu2025agentsentinelendtoendrealtimesecurity}.

\noindent\textbf{Metrics.}
For safety, we report Attack Success Rate (ASR$\downarrow$) on ASB and AgentDojo, and Safety Score$\uparrow$ on Agent-SafetyBench. For utility, we report Performance under No Attack (PNA$\uparrow$) on ASB and Utility$\uparrow$ on AgentDojo. On ASB, we additionally report Net Resilient Performance (NRP$\uparrow$) as the combined capability, calculated by $\mathrm{NRP}=\mathrm{PNA}\times(1-\mathrm{ASR})$. Since defenses primarily target safety, we emphasize safety-oriented metrics while reporting utility to verify benign-task preservation.

More details about experimental hardware, software, and hyperparameters are in App.~\ref{app:details} and App.~\ref{app:sampling-size}.
% App.~\ref{app:details} and App.~\ref{app:sampling-size}.

% \tzlconfirmed{\noindent\textbf{Hyper-parameters.}
% We set $\alpha=0.95,\mathcal{B}=20,K=200,\sigma(\cdot)=e^{(\cdot)}$.
% % \gamma_1=0.4,\gamma_2=0.6$. 
% We choose $\alpha=0.95,\mathcal{B}=20$ based on analysis on utility-safety tradeoff in Sec.~\ref{4.4}, and $K=200$ based on a sampling-size trade-off analysis between ASR and computational cost in App.~\ref{app:sampling-size}. Other implementation details are shown in App.~\ref{app:details}.}

% \noindent\textbf{Computational overhead.}
% We report the overall computation overhead of different defense methods and a component-wise breakdown of TAS in App.~\ref{app:compute_overhead}. 

\subsection{Overall Performance}

\label{4.2}

\begin{table*}[t]
\centering
\footnotesize
\setlength{\tabcolsep}{1.2pt}
\renewcommand{\arraystretch}{0.90}

% Compact rule spacing, local to this table only.
\begingroup
\setlength{\aboverulesep}{0.8pt}
\setlength{\belowrulesep}{0.8pt}
\setlength{\cmidrulesep}{0.7pt}

\begin{tabular*}{\textwidth}{
    @{\extracolsep{\fill}}
    l l
    *{12}{c}
    @{}
}
\toprule
\multirow{2}{*}{\textbf{Model}}
&
\multirow{2}{*}{\textbf{Metric}}
&
\multicolumn{6}{c}{\textbf{DPI}}
&
\multicolumn{6}{c}{\textbf{IPI}}
\\
\cmidrule(lr){3-8}
\cmidrule(lr){9-14}
&
&
Ours
&
w/o MDP
&
w/o Tail
&
w/o SRM
&
w/o Constr.
&
w/o Samp.
&
Ours
&
w/o MDP
&
w/o Tail
&
w/o SRM
&
w/o Constr.
&
w/o Samp.
\\
\midrule

\multirow{3}{*}{Llama3-70B}
& ASR $\downarrow$
& \textbf{43.50}
& 50.75
& 69.75
& \underline{50.50}
& 87.75
& 66.25
& \textbf{13.95}
& \underline{18.50}
& 22.25
& 18.75
& 45.50
& 26.00
\\
& NRP $\uparrow$
& \textbf{32.91}
& \underline{27.21}
& 14.97
& 23.64
& 6.62
& 18.39
& \textbf{50.12}
& \underline{45.03}
& 38.49
& 38.80
& 29.43
& 40.33
\\
& PNA $\uparrow$
& 58.25
& 55.25
& 49.50
& 47.75
& 54.00
& 54.50
& 58.25
& 55.25
& 49.50
& 47.75
& 54.00
& 54.50
\\
\midrule

\multirow{3}{*}{Qwen2-72B}
& ASR $\downarrow$
& \textbf{48.35}
& 70.00
& 96.50
& \underline{69.00}
& 99.75
& 94.75
& \textbf{17.05}
& 22.25
& 23.50
& \underline{21.25}
& 27.25
& 26.75
\\
& NRP $\uparrow$
& \textbf{11.36}
& \underline{6.68}
& 0.78
& 6.28
& 0.06
& 1.18
& \textbf{18.25}
& \underline{17.30}
& 17.02
& 15.95
& 16.73
& 16.48
\\
& PNA $\uparrow$
& 22.00
& 22.25
& 22.25
& 20.25
& 23.00
& 22.25
& 22.00
& 22.25
& 22.25
& 20.25
& 23.00
& 22.50
\\
\midrule

\multirow{3}{*}{GPT-4o}
& ASR $\downarrow$
& \textbf{40.35}
& \underline{54.45}
& 71.75
& 67.50
& 73.25
& 68.50
& \textbf{45.65}
& 60.25
& 72.50
& \underline{59.75}
& 73.75
& 66.75
\\
& NRP $\uparrow$
& \textbf{47.42}
& \underline{36.10}
& 20.41
& 20.72
& 21.40
& 25.04
& \textbf{43.21}
& \underline{31.50}
& 19.87
& 25.66
& 21.00
& 26.43
\\
& PNA $\uparrow$
& 79.50
& 79.25
& 72.25
& 63.75
& 80.00
& 79.50
& 79.50
& 79.25
& 72.25
& 63.75
& 80.00
& 79.50
\\
\midrule

\multirow{3}{*}{GPT-5.2}
& ASR $\downarrow$
& \textbf{48.40}
& \underline{53.75}
& 77.75
& 63.00
& 81.25
& 72.35
& \textbf{42.25}
& \underline{50.00}
& 68.75
& 55.00
& 73.50
& 61.75
\\
& NRP $\uparrow$
& \textbf{48.89}
& \underline{43.71}
& 20.41
& 32.93
& 17.72
& 26.06
& \textbf{54.72}
& \underline{47.25}
& 28.67
& 40.05
& 25.04
& 36.05
\\
& PNA $\uparrow$
& 94.75
& 94.50
& 91.75
& 89.00
& 94.50
& 94.25
& 94.75
& 94.50
& 91.75
& 89.00
& 94.50
& 94.25
\\
\bottomrule
\end{tabular*}

\endgroup

\caption{
Ablation results for TAS under DPI and IPI attacks.
``w/o'' denotes removing the corresponding component.
}
\label{tab:ablation}
\end{table*}

Tab.~\ref{tab:main_results} reports ASB results across four LLM backbones. Our TAS consistently improves safety and combined capability while keeping utility close to baselines. Compared with \emph{No Defense}, TAS achieves 47.95\% and 70.25\% average safety improvements under DPI and IPI attacks, respectively, and further outperforms previous SOTA baselines by 24.28\% and 20.53\%. TAS also obtains the best average NRP under both attacks, indicating a better balance between benign-task utility and attack resistance.\footnote{We adopt the same execution prompt as baselines when our budget is satisfied, for fair comparisons.}
Tab.~\ref{tab:main_pot} reports ASR results under POT backdoor attacks. TAS achieves the lowest average ASR across four LLM backbones, reducing it by 58.69\% compared with \emph{No Defense} and by 44.64\% compared with the strongest POT baseline. Full utility and NRP results are provided in App.~\ref{app:exp_pot}.
% App.~\ref{app:exp_pot}.
% 
Tab.~\ref{tab:combined-benchmark-results} further evaluates our TAS on Agent-SafetyBench and AgentDojo against recent agent-safety baselines. On Agent-SafetyBench, TAS achieves the best average Safety Score, improving 61.36\% over \emph{No Defense} and 11.41\% over the strongest baseline. On AgentDojo, TAS achieves the lowest ASR, reducing GPT-4o ASR by up to 91.73\% compared with \emph{No Defense}, while maintaining competitive utility. These results show that TAS generalizes beyond ASB to broader safety-oriented and tool-use agent benchmarks. 
Notably, experiments across both open-source and closed-source LLMs show the effectiveness of TAS, including strong performance on the thinking model GPT-5.2.

% \noindent\textbf{More experimental analysis.}
% % We report computation overhead in App.~\ref{app:compute_overhead} and provide tail-risk visualization analysis in App.~\ref{app:exp_analysis_tail}, supporting our motivation that TAS suppresses rare high-cost failures.
% We report computation overhead in App.~\ref{app:compute_overhead}.
% We further analyze the reliability of TAS risk evaluation in App.~\ref{app:evaluator_robustness} and App.~\ref{app:fpr_fnr}. Under $10\%$ evaluator score misclassifications, TAS slightly causes about a 1\% relative safety drop. TAS also achieves both a low false positive rate and false negative rate, indicating balanced risk evaluation with fewer unnecessary interventions and missed risks. This robustness comes from estimating the risk distribution via multiple samplings (\S~\ref{3.4}) rather than relying on a single evaluator score.

\subsection{Ablation Studies}
\label{4.3}

As shown in Tab.~\ref{tab:ablation}, we conduct ablation studies to validate the contribution of each component in TAS. 
Removing MDP modeling (w/o mdp) ignores action-conditioned risk propagation and leads to higher ASR, showing the importance of modeling sequential dependencies. 
Replacing the tail-sensitive measure with an overall average (w/o tail) causes clear safety and utility drops, confirming that average-based risk measurement can miss rare high-cost failures. 
Removing SRM weighting (w/o srm, $\sigma$ in Eq.~\eqref{eq:risk_score}) also worsens safety compared with the full method, indicating the benefit of assigning larger weights to higher-risk tail samples in SRM. 
Removing the shared budget constraint (w/o constraint) causes the largest safety degradation, showing that the budget is critical for controlling risk propagation. 
Finally, replacing empirical estimation with a single sample (w/o sampling, Eq.~\eqref{eq:emp_cost}) leads to poorer safety, demonstrating the importance of sampling-based risk estimation.

% We further analyze the robustness of TAS to evaluator score misclassifications in App.~\ref{app:evaluator_robustness}, where introducing $10\%$ score misclassifications only slightly causes about a 1\% relative safety drop, indicating the robustness and reliability of our evaluator. That is because we take the risk distribution by multiple samplings (\S~\ref{3.4}) rather than a single score for robustness. We additionally report false positive and false negative rates in App.~\ref{app:fpr_fnr} to evaluate the reliability of risk evaluation.

% We further analyze the reliability of TAS risk evaluation in App.~\ref{app:evaluator_robustness} and App.~\ref{app:fpr_fnr}. Under $10\%$ evaluator score misclassifications, TAS slightly causes about a 1\% relative safety drop. TAS also achieves both a low false positive rate and false negative rate, indicating balanced risk evaluation with fewer unnecessary interventions and missed risks. This robustness comes from estimating the risk distribution via multiple samplings (\S~\ref{3.4}) rather than relying on a single evaluator score.

% \subsection{Analysis Studies}
% \subsection{Tail Risk Visualization}

% \subsection{Analysis on Tail Risk}

\subsection{Analysis on Tail-risk Suppression}
% \paragraph{Analysis on Tail-risk Suppression.}
\label{4.5}

% \tzlconfirmed{To verify the motivation of whether our method actually constrains tail risks in agent workflows, we conduct a tail-risk visualization analysis comparing our method to other baselines.
% Specifically, we run the same workflow with our method and the baselines to record the risk costs at each step using the same risk scoring.
% We aggregate risk costs from each run and plot the tail quantile curve, in which the x-axis is the tail probability (smaller means rarer), and the y-axis is the risk cost threshold. A point $(x, y)$ means that a fraction $x$ of steps have risk costs more than $y$. Therefore, the closer to the upper left in the figure, the larger the tail risk the point has. }
% We expect baselines to a sharp rise towards the upper left, while our method stays at the lower left.

% To verify whether our method constrains tail risks in agent workflows, we compare tail quantile curves between our method and baselines. We run the same workflow under each method, record step-wise risk costs with the same risk scorer, and plot the risk-cost threshold against the tail probability. Points closer to the upper left indicate larger rare-event risks.

To verify whether our method constrains tail risks in agent workflows, we conduct a tail-risk visualization analysis against baselines. Specifically, we run the same workflow under each method, record step-wise risk costs using the same risk scorer, and aggregate them into tail quantile curves. In the curve, the x-axis denotes the tail probability, and the y-axis denotes the corresponding risk-cost threshold. Points closer to the upper left indicate that high-risk costs occur at small tail probabilities.

\begin{figure*}[t]
    \centering
    \setlength{\tabcolsep}{1pt}
    \begin{tabular}{cccc}
        \begin{subfigure}[t]{0.2425\textwidth}
            \centering
            \includegraphics[width=\linewidth]{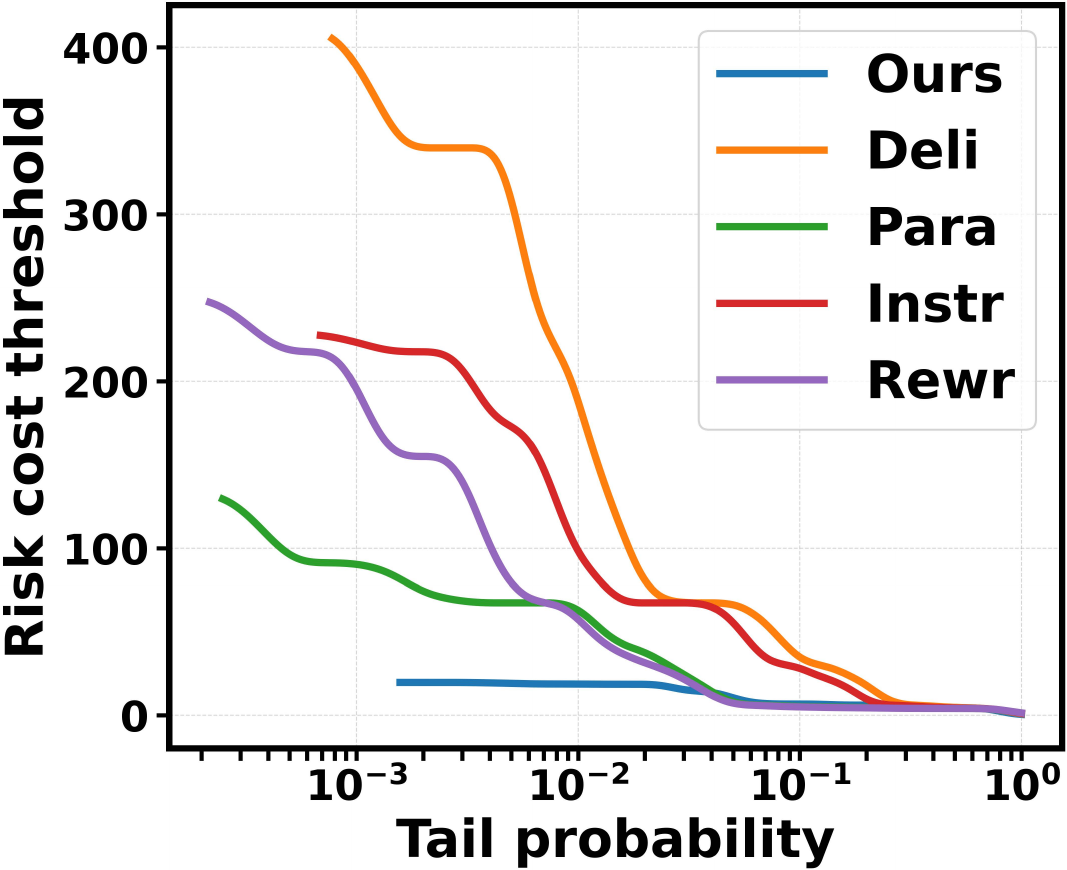}
            \caption{LlaMA3-70B}
            \label{fig:tail_llama3_70b}
        \end{subfigure}
        &
        \begin{subfigure}[t]{0.2425\textwidth}
            \centering
            \includegraphics[width=\linewidth]{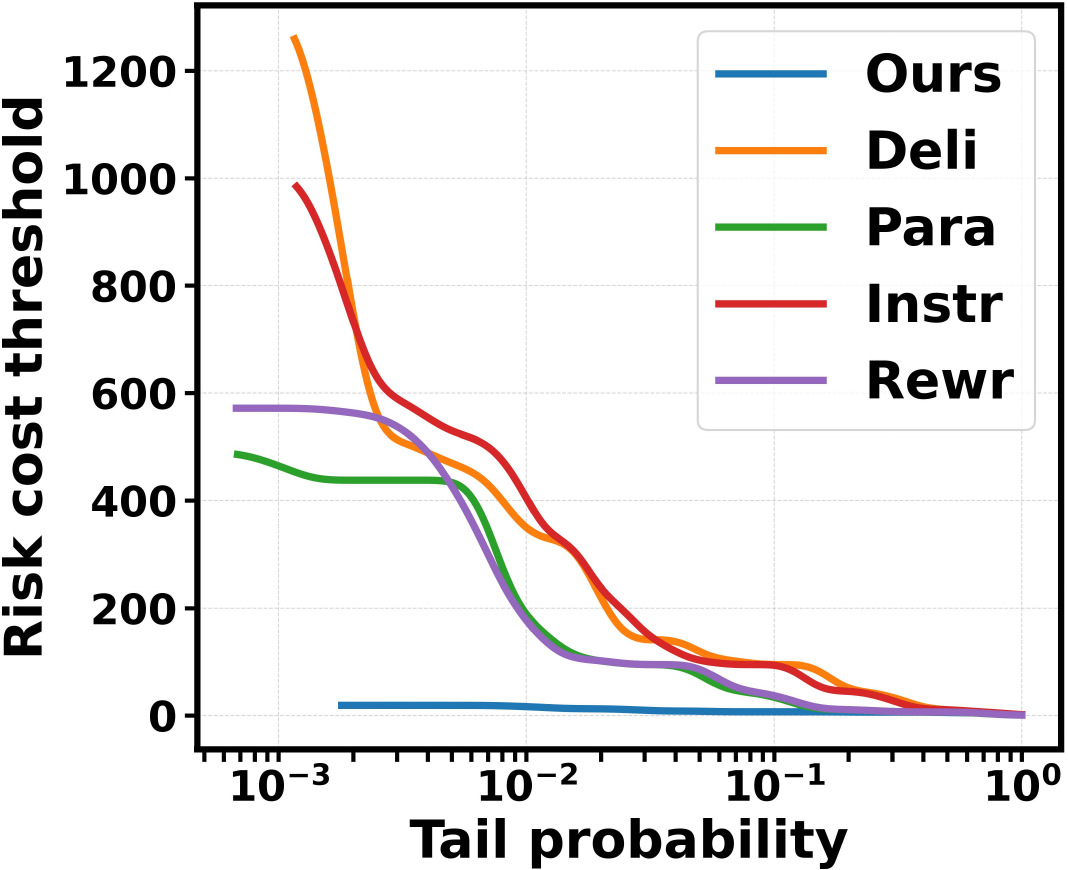}
            \caption{Qwen2-72B}
            \label{fig:tail_qwen2_72b}
        \end{subfigure}
        &
        \begin{subfigure}[t]{0.2425\textwidth}
            \centering
            \includegraphics[width=\linewidth]{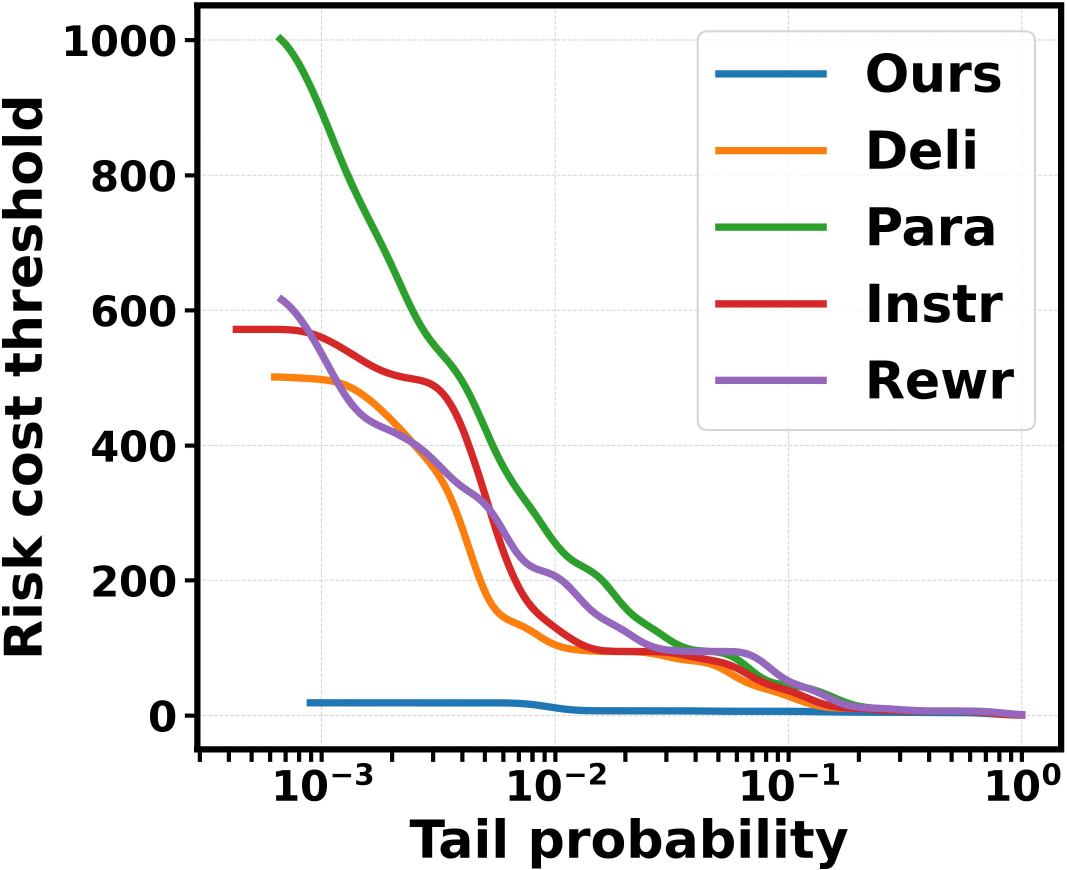}
            \caption{GPT-4o}
            \label{fig:tail_gpt4o}
        \end{subfigure}
        &
        \begin{subfigure}[t]{0.2425\textwidth}
            \centering
            \includegraphics[width=\linewidth]{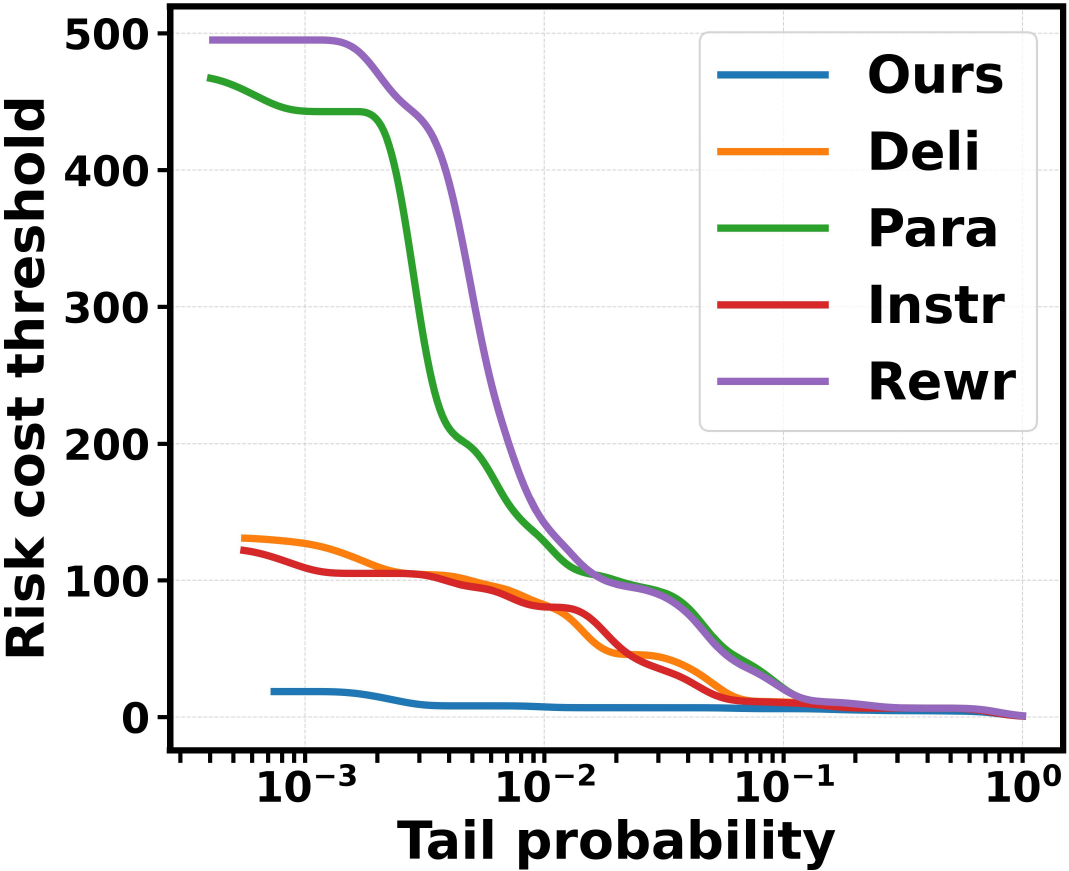}
            \caption{GPT-5.2}
            \label{fig:tail_gpt52}
        \end{subfigure}
    \end{tabular}

    \caption{\tzlconfirmed{Tail-risk visualization of our method and four baselines (curves with different colors) under DPI attack across four LLM backbones (Fig.~\ref{fig:tail_llama3_70b}, \ref{fig:tail_qwen2_72b}, \ref{fig:tail_gpt4o}, and \ref{fig:tail_gpt52}). The X-axis indicates the tail probability, and the Y-axis indicates the risk cost threshold. Lower x-value and higher y-value indicate high cost with low-probability (i.e., tail risk).}}
    \vspace{-10pt}
    \label{fig:tail_four_models}
\end{figure*}

\tzlconfirmed{As shown in Fig.~\ref{fig:tail_four_models}, we find the same trend holding across the four LLM backbones, where baseline curves present a sharp rise towards the upper left, while our method consistently stays at the lower left,
% our curve stays consistently lower than baselines in rare probability (e.g., $x\le10^{-2}$), 
indicating that our method has substantially smaller risks in the tails.
In contrast, baseline curves increase steeply near the left end, showing an apparent tail risk.
The results confirm that our method effectively suppresses tail risk to ensure better safety in the workflow.}

\subsection{Analysis on Utility-safety Tradeoff}

% \paragraph{Analysis on Utility-safety Tradeoff.}
\label{4.4}

% aaai单栏
\begin{figure}[t]
    \centering

    \begin{subfigure}[t]{0.48\columnwidth}
        \centering
        \includegraphics[width=\linewidth]{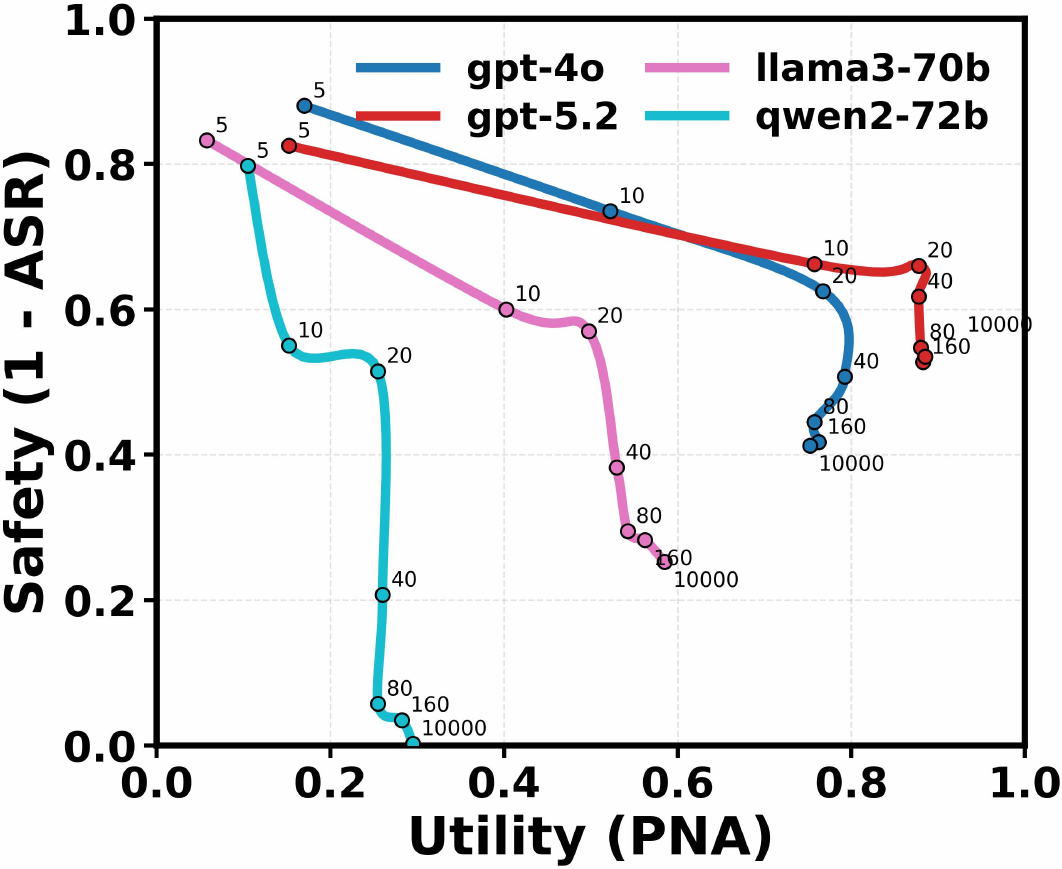}
        \caption{Budget $\mathcal{B}$}
        \label{fig:tradeoff_budget}
    \end{subfigure}
    \hfill
    \begin{subfigure}[t]{0.48\columnwidth}
        \centering
        \includegraphics[width=\linewidth]{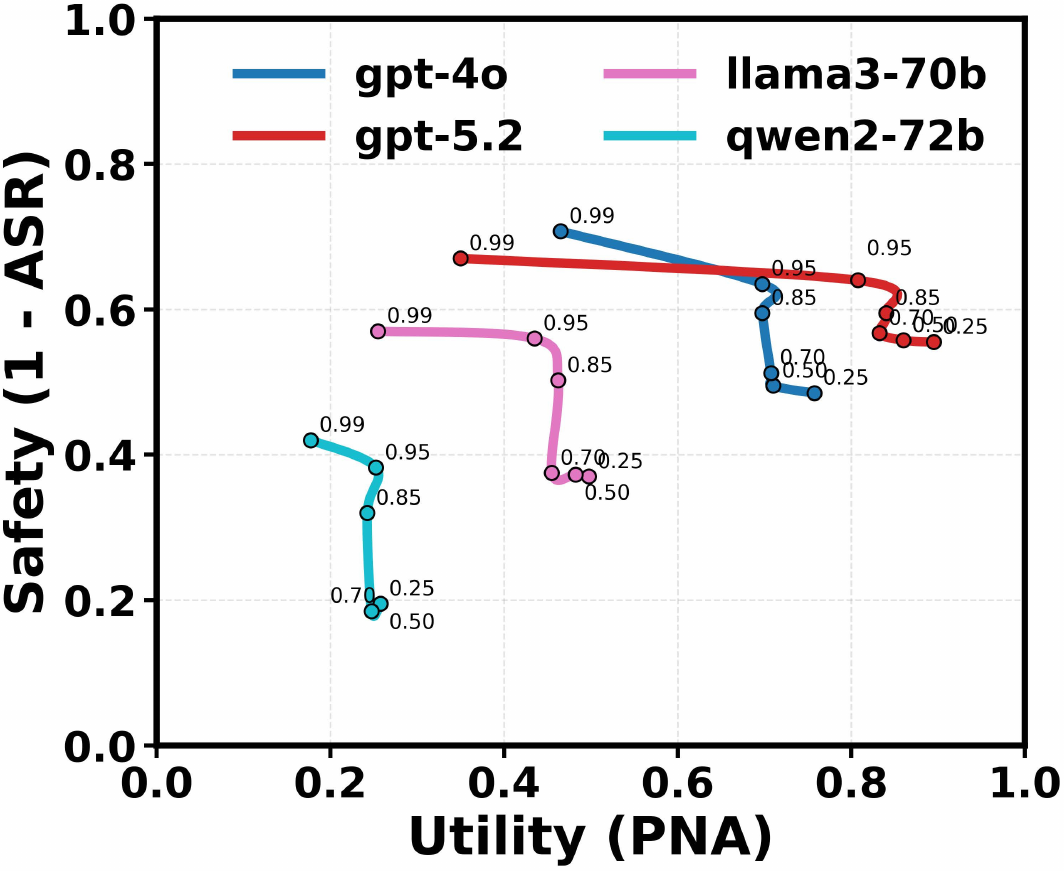}
        \caption{Quantile cutoff $\alpha$}
        \label{fig:tradeoff_alpha}
    \end{subfigure}

    \caption{
    Utility-safety trade-offs under DPI attacks with varying
    budgets and quantile cutoffs across four LLM backbones.
    Higher values indicate better utility and safety.
    }
    \label{fig:tradeoff_all}
\end{figure}

To explore the tradeoff between utility and safety in our method, we conduct an analysis experiment by varying the dynamic budget $\mathcal{B}$ in Eq.~\eqref{eq:dynamic_budget} and the quantile cutoff $\alpha$ in Eq.~\eqref{eq:risk_score}. We separately adjust one parameter while keeping the other fixed, and evaluate how these two factors affect the balance between task utility and safety. Detailed parameter settings are provided in App.~\ref{app:details}.

As shown in Fig.~\ref{fig:tradeoff_all}, the same trend holds across four LLM backbones: increasing $\mathcal{B}$ or decreasing $\alpha$ improves utility but reduces safety.
We analyze the results from two aspects:
\textbf{(1) Budgets.} Smaller budgets impose stricter constraints, which better block high-risk steps but may over-reject benign executions. As $\mathcal{B}$ increases, more benign steps pass, and utility improves, but more high-risk steps are also allowed, leading to safety degradation.
\textbf{(2) Quantile Cutoffs.} Smaller $\alpha$ considers a larger tail region, weakening the emphasis on extreme risks. This improves utility by allowing more steps to pass, but reduces safety by under-penalizing rare high-risk cases.
Overall, $\mathcal{B}\approx20$ and $\alpha\approx0.95$ provide a practical sweet point for balancing utility and safety.

\noindent\textbf{Additional analyses and comparisons.}
We provide additional attack and guardrail comparisons, computation-overhead results, and risk-evaluation reliability analyses in App.~\ref{app:exp_pot} and App.~\ref{app:exp_guard}, demonstrating that TAS remains effective beyond the main settings while being robust to $10\%$ evaluator score misclassification and maintaining low false positive and false negative rates.

% \paragraph{Additional analyses and comparisons.}
% We provide supplementary comparisons under additional attacks and original guardrail evaluation protocols in App.~\ref{app:exp_pot} and App.~\ref{app:exp_guard}, showing that TAS remains effective beyond the main settings. 
% We report computation overhead in App.~\ref{app:compute_overhead}. 
% We further analyze risk-evaluation reliability in App.~\ref{app:evaluator_robustness} and App.~\ref{app:fpr_fnr}, showing robustness to $10\%$ evaluator score misclassification and low false positive/negative rates.

\section{Conclusion}
\tzlconfirmed{
In this paper, we propose TAS to improve safety for multi-step LLM-based agent workflows.
% Instead of a single-step safety check, TAS models the workflow safety with MDP to trace dependent safety state transitions, then applies a shared budget to explicitly constrain tail risk, and finally optimizes the workflow under the constraint to provide a safety strategy for the current step.
% 
% Instead of single-step safety checks, TAS models the workflow as an MDP over dependent safety states, measures tail risk with spectral risk, and applies a shared budget with empirical estimates to make a step-wise safety strategy.
% 
Instead of single-step safety checks, TAS models the workflow as an MDP over dependent safety states to capture action-conditioned risk propagation. It then measures tail risk with spectral risk and empirically estimates the tail risk distribution by sampling. Constrained on a shared budget, TAS makes a step-wise safety strategy to suppress rare high-cost failures while preserving utility.
Our main results on three metrics show that TAS consistently outperforms baselines across six LLM backbones.
%Ablations confirm that each component contributes to performance.
Analyses of tail risk and the utility-safety tradeoff show that TAS suppresses rare high-cost risks while identifying practical sweet points that balance safety and utility.}
% The ablation studies in \S~\ref{4.3} verify the effectiveness of different parts of our method.
% We visualize the tail risk of our method and baselines in \S~\ref{4.4} to prove that our motivation for tail risk measures works.
% We also explore the tradeoff between utility and safety with different budgets and quantiles in \S~\ref{4.5}, finding the best budget and quantile points to balance the utility-safety tradeoff.
% 
% We further discuss the societal impacts and limitations of our methods in App.~\ref{app:impacts_limitations}.

% \small\bibliography{aaai2027}

\bibliography{aaai2027}
\clearpage
\appendix

% \section{Supplementary Experimental Details}

\section{Implementation Details}
\label{app:details}
We provide more implementation details for our experimental settings as shown in Tab.~\ref{tab:impl-details}. 
In Sec.~{4.4}, for the budget analysis, we test $\mathcal{B}\in\{5,10,20,40,80,160,10000\}$ with fixed $\alpha=0.95$. 
For the quantile cutoff analysis, we test $\alpha\in\{0.99,0.95,0.85,0.70,0.50,0.25\}$ with fixed $\mathcal{B}=20$. 
We choose $\alpha=0.95,\mathcal{B}=20$ based on analysis on utility-safety tradeoff in Sec.~{4.4}, and $K=200$ based on a sampling-size trade-off analysis between ASR and computational cost.
% in App.~\ref{app:sampling-size}.
% Appendix: Implementation Details
% \begin{table}[h]
%   \centering
%   \setlength{\tabcolsep}{32pt}
%   \renewcommand{\arraystretch}{1.15}
%   \caption{Implementation details of our experimental environment.}
%   \begin{tabular}{lc}
%   % {p{0.28\linewidth} >{\centering\arraybackslash}p{0.66\linewidth}}
%     \toprule
%     \textbf{Item} & \textbf{Setting} \\
%     \midrule
%     OS & Ubuntu 22.04 LTS \\
%     GPU & 1$\times$ NVIDIA A100 80GB PCIe \\
%     NVIDIA Driver Version & 560.35.03 \\
%     CPU Architecture & x86\_64 \\
%     CPU Model & Intel(R) Xeon(R) Platinum 8358 @ 2.60GHz \\
%     CPU Threads & 128 \\
%     Memory & 1T \\
%     Python & 3.10.19 \\
%     PyTorch & 2.5.1+cu121 \\
%     Ollama Version & 0.13.3 \\
%     LLM Quantization & q4\_K\_M \\
%     \midrule
%     $\alpha$ & 0.95 \\
%     $\mathcal{B}$ & 20 \\
%     $K$ & 200 \\
%     $\sigma(\cdot)$ & $e^{(\cdot)}$ \\
%     $\gamma_1,\gamma_2$ & 0.4, 0.6 \\
%     API score model & gpt-5-mini \\
%     \bottomrule
%   \end{tabular}
%   \label{tab:impl-details}
% \end{table}

\begin{table}[h]
  \centering
  \setlength{\tabcolsep}{4pt}
  \renewcommand{\arraystretch}{1.15}
  \begin{tabularx}{\columnwidth}{@{}l >{\raggedleft\arraybackslash}X@{}}
    \toprule
    \textbf{Item} & \textbf{Setting} \\
    \midrule
    OS & Ubuntu 22.04 LTS \\
    GPU & 12$\times$ NVIDIA A100 80GB PCIe \\
    NVIDIA Driver Version & 560.35.03 \\
    CPU Architecture & x86\_64 \\
    CPU Model & Intel(R) Xeon(R) Platinum 8358 @ 2.60GHz \\
    CPU Threads & 128 \\
    Memory & 1T \\
    Python & 3.10.19 \\
    PyTorch & 2.5.1+cu121 \\
    Ollama Version & 0.13.3 \\
    vLLM Version & 0.7.3 \\
    LLM Quantization & q4\_K\_M \\
    % Tensor Parallel Size & 2 GPUs per vLLM worker replica \\
    % Number of vLLM Worker Replicas & 6 \\
    Batch Size & 256 \\
    \midrule
    $\alpha$ & 0.95 \\
    $\mathcal{B}$ & 20 \\
    $K$ & 200 \\
    $\sigma(\cdot)$ & $e^{(\cdot)}$ \\
    % $\gamma_1,\gamma_2$ & 0.4, 0.6 \\
    % Max Retries $R$ & 3 \\
    Sampling Temperature & 0.7 \\
    Sampling Top-p & 0.95 \\
    API Score Model & gpt-5-mini \\
    \bottomrule
  \end{tabularx}
  \caption{Implementation details of our experiments.}
  \label{tab:impl-details}
\end{table}

% \section{Computation Overhead}
% \label{app:compute_overhead}

% As shown in Tab.~\ref{tab:compute_overhead}, we report computation overheads of different baselines and our methods, relatively compared to \emph{No Defense}. 
% The results show that our method requires approximately 34\% more of the computational time compared to \emph{No Defense} because of sampling for estimation of tail risk distribution in \S~\ref{3.4}.

% \input{tables/computation_cost}

% To further identify the source of the overhead, we provide a component-wise breakdown in Tab.~\ref{tab:overhead_breakdown}. As for MDP, it's our theoretical modeling, which does not practically introduce additional latency. As for the evaluator, it additionally introduces 14.7\% relative latency.

% \input{tables/component_wise_overhead}

\section{Supplementary Experiments}
\label{app:exp_all}

\subsection{Additional ASB and Ablation Results of Smaller LLMs}
\label{app:exp_full_main}

We also report additional ASB and ablation results of the smaller LLMs in the following Tab.~\ref{tab:app_main_results} and Tab.~\ref{tab:app_ablation}, respectively.

\begin{table*}[h]
\centering
\small
\setlength{\tabcolsep}{5pt}
\renewcommand{\arraystretch}{1.15}
\resizebox{\textwidth}{!}{%
\begin{tabular}{@{}clccccccccccc@{}}
\toprule
\multirow{2}{*}{Model} & \multirow{2}{*}{Metric} & \multicolumn{6}{c}{DPI} & \multicolumn{5}{c}{IPI} \\
\cmidrule(lr){3-8}\cmidrule(lr){9-13}
 &  & No Defense & Deli. & Para. & Instr. & Rewr. & Ours & No Defense & Deli. & Instr. & Sand. & Ours \\
\midrule

\multirow{3}{*}{LlaMA3-8B} & ASR & 33.75 & 62.75 & 28.50 & 52.00 & \underline{23.15} & \textbf{22.50} & 11.50 & 9.25 & \underline{8.75} & 13.00 & \textbf{2.85} \\
 & NRP & 0.99 & 0.65 & 1.79 & \underline{2.40} & 1.92 & \textbf{3.68} & 1.33 & 1.59 & \underline{4.56} & 2.61 & \textbf{4.61} \\
  & PNA & 1.50 & 1.75 & 2.50 & {5.00} & 2.50 & {4.75} & 1.50 & 1.75 & {5.00} & 3.00 & {4.75} \\
\midrule

% \multirow{3}{*}{LlaMA3-70B} & ASR & 87.75 & 88.25 & \underline{71.25} & 87.25 & 74.50 & \textbf{43.50} & 45.50 & \underline{34.50} & 41.50 & 39.75 & \textbf{13.95} \\
%  & NRP & 6.80 & 6.76 & \underline{14.81} & 7.87 & 12.69 & \textbf{32.91} & 30.25 & \underline{37.66} & 36.12 & 32.54 & \textbf{50.12} \\
%   & PNA & 55.50 & 57.50 & 51.50 & {61.75} & 49.75 & {58.25} & 55.50 & 57.50 & {61.75} & 54.00 & {58.25} \\
% \midrule

\multirow{3}{*}{Qwen2-7B} & ASR & 73.50 & 80.00 & 46.25 & 76.75 & \underline{33.30} & \textbf{12.80} & \underline{9.25} & 11.25 & 9.50 & 11.00 & \textbf{1.65} \\
 & NRP & 0.60 & 0.40 & 0.94 & 0.76 & \underline{2.00} & \textbf{3.05} & 2.04 & 1.77 & \underline{2.94} & 2.00 & \textbf{3.44} \\
  & PNA & 2.25 & 2.00 & 1.75 & {3.25} & 3.00 & {3.50} & 2.25 & 2.00 & {3.25} & 2.25 & {3.50} \\
\bottomrule
\end{tabular}
}
\caption{\tzlconfirmed{Additional ASB results across smaller LLM backbones. 
% We report the computation overheads of baselines and our methods in App.~\ref{app:compute_overhead}.
}}
% \vspace{-10pt}
\label{tab:app_main_results}
\end{table*}

\begin{table*}[h]
\centering
\small
\setlength{\tabcolsep}{1.2pt}
\renewcommand{\arraystretch}{1.08}
\resizebox{\textwidth}{!}{%
\begin{tabular}{@{}llcccccccccccc@{}}
\toprule
\multirow{2}{*}{Model} & \multirow{2}{*}{Metric}
& \multicolumn{6}{c}{DPI}
& \multicolumn{6}{c}{IPI} \\
\cmidrule(lr){3-8} \cmidrule(lr){9-14}
& & Ours & w/o mdp & w/o tail & w/o srm
& \makecell{w/o\\constraint} & \makecell{w/o\\sampling}
& Ours & w/o mdp & w/o tail & w/o srm
& \makecell{w/o\\constraint} & \makecell{w/o\\sampling} \\
\midrule

\multirow{3}{*}{LlaMA3-8B} & ASR & \textbf{22.50} & \underline{27.25} & 37.75 & 31.50 & {38.78} &38.00& \textbf{2.85} & 3.25 & 9.50 & 4.50 & {11.50} &\underline{3.00} \\
 & NRP & \textbf{3.68} & \underline{3.27} & 1.40 & 1.20 & 2.76 & 2.79 & \textbf{4.61} & 4.35 & 2.04 & 1.67 &3.98 &\underline{4.37} \\
 & PNA & {4.75} & 4.50 & 2.25 & 1.75 & {4.50} & 4.50 & {4.75} & 4.50 & 2.25 & 1.75 & {4.50} &4.50 \\
\midrule

% \multirow{3}{*}{LlaMA3-70B} & ASR & \textbf{43.50} & 50.75 & 69.75 & \underline{50.50} &87.75 & {66.25} & \textbf{13.95} & \underline{18.50} & 22.25 & 18.75 & 45.50& {26.00} \\
%  & NRP & \textbf{31.36} & \underline{27.21} & 14.97 & {23.64} &6.62 & 18.39 & \textbf{47.76} & \underline{45.03} & 38.49 & {38.80} & 29.43 &40.33\\
%  & PNA & {55.50} & {55.25} & 49.50 & 47.75 & 54.00 &54.50 & {55.50} & {55.25} & 49.50 & 47.75 & 54.00 &54.50 \\
% \midrule

\multirow{3}{*}{Qwen2-7B} & ASR & \textbf{12.80} & \underline{14.50} & 22.00 & 17.00 & {73.50} &24.65 & \textbf{1.65} & 2.50 & 3.00 & \underline{2.00} & {9.25} &6.25\\
 & NRP & \textbf{3.05} & \underline{2.78} & {2.15} & 1.25 & 0.93 &2.45 & \textbf{3.44} & {3.17} & 2.67 & 1.47 & \underline{3.18} &3.05 \\
 & PNA & {3.50} & {3.25} & 2.75 & 1.50 & 3.50 & {3.25} & {3.50} & {3.00} & 2.75 & 1.50 & 3.50 & {3.25} \\
% \midrule

% \multirow{3}{*}{Qwen2-72B} & ASR & \textbf{48.35} & 70.00 & 96.50 & \underline{69.00} & {99.75} & 94.75& \textbf{17.05} & 22.25 & 23.50 & \underline{21.25} & 27.25& {26.75} \\
%  & NRP & \textbf{11.36} & \underline{6.68} & 0.78 & 6.28 & {0.06} &1.18 & \textbf{18.25} & \underline{17.30} & 17.02 & 15.95 & 16.73 &16.48 \\
%  & PNA & {22.00} & {22.25} & {22.25} & 20.25 & {23.00}& {22.25} & 22.00 & {22.25} & {22.25} & 20.25 & {23.00} &{22.50} \\
% \midrule

% \multirow{3}{*}{GPT-4o} & ASR & \textbf{40.35}& \underline{54.45}& 71.75& 67.50& {73.25} & 68.50& \textbf{45.65}& 60.25& 72.50& \underline{59.75}& {73.75} &66.75 \\
%  & NRP & \textbf{47.42}& \underline{36.10}& 20.41& 20.72&21.40 & {25.04}& \textbf{43.21}& \underline{31.50}& 19.87& 25.66 & 21.00 &{26.43} \\
%  & PNA & {79.50}& {79.25}& 72.25& 63.75& {80.00}& 79.50 & {79.50}& {79.25}& 72.25& 63.75& {80.00} &79.50 \\
% \midrule

% \multirow{3}{*}{GPT-5.2} & ASR & \textbf{48.40}& \underline{53.75}& 77.75& 63.00& {81.25} & 72.35 & \textbf{42.25}& \underline{50.00}& 68.75& 55.00 & 73.50& {61.75} \\
%  & NRP & \textbf{48.89}& \underline{43.71}& 20.41& 32.93& {17.72} & 26.06& \textbf{54.72}& \underline{47.25}& 28.67& 40.05& {25.04} &36.05 \\
%  & PNA & {94.75}& 94.50& 91.75& 89.00& {94.50}& {94.25}& {94.75} & 94.50 & 91.75& 89.00& {94.50} & 94.25 \\

\bottomrule
\end{tabular}
}
\caption{\tzlconfirmed{Ablations on the small LLMs for different parts of our method. W/o means the removal of the corresponding part.}}
\label{tab:app_ablation}
\end{table*}

\subsection{Complete Results under the POT Attack}
\label{app:exp_pot}
% As shown in Tab.~\ref{tab:app_pot}, keeping the same ASB benchmark \cite{zhang2025agent} as in our main experiments, we extend defense comparisons under POT backdoor attacks \cite{xiang2024badchain}, and compare our method to two baseline defenses (i.e., \texttt{Shuffle} \cite{xiang2023cbd} and \texttt{Para.}). The reason why the baselines are different from our main experiments is that different defense baselines are suitable for different attack scenarios, while our method is applicable under various attacks.
% The results show that our method consistently achieves better safety than baselines and also maintains the close utility as the \emph{No defense}.
Tab.~\ref{tab:app_pot} provides the complete POT results. We use the same ASB benchmark~\cite{zhang2025agent} and compare TAS with two POT-specific baseline defenses, \texttt{Para.} and \texttt{Shuffle}~\cite{xiang2023cbd}. The baselines differ from those in the DPI/IPI experiments because different defense methods are suitable for different attack settings, while TAS is applicable across attacks.
The results show that TAS not only achieves lower ASR, but also obtains the best average NRP while maintaining PNA close to \emph{No Defense}, demonstrating a favorable safety-utility tradeoff under POT attacks.

\begin{table*}[t]
\centering
\small
\setlength{\tabcolsep}{3pt}
\renewcommand{\arraystretch}{0.92}

\begingroup
\setlength{\aboverulesep}{0.8pt}
\setlength{\belowrulesep}{0.8pt}

\begin{tabular*}{0.82\textwidth}{
    @{\extracolsep{\fill}}
    l l cccc
    @{}
}
\toprule
\textbf{Model}
&
\textbf{Metric}
&
\textbf{No Def.}
&
\textbf{Shuffle}
&
\textbf{Para.}
&
\textbf{Ours}
\\
\midrule

\multirow{3}{*}{Llama3-70B}
& ASR $\downarrow$
& 57.00
& 63.75
& \underline{44.75}
& \textbf{15.00}
\\
& NRP $\uparrow$
& \underline{23.87}
& 16.04
& 23.48
& \textbf{49.51}
\\
& PNA $\uparrow$
& 55.50
& 44.25
& 42.50
& 58.25
\\
\midrule

\multirow{3}{*}{Qwen2-72B}
& ASR $\downarrow$
& 96.50
& 81.50
& \underline{57.75}
& \textbf{43.25}
\\
& NRP $\uparrow$
& 0.69
& 4.30
& \underline{10.03}
& \textbf{12.49}
\\
& PNA $\uparrow$
& 19.75
& 23.25
& 23.75
& 22.00
\\
\midrule

\multirow{3}{*}{GPT-4o}
& ASR $\downarrow$
& 100.00
& 98.50
& \underline{84.75}
& \textbf{42.75}
\\
& NRP $\uparrow$
& 0.00
& 1.16
& \underline{11.70}
& \textbf{45.51}
\\
& PNA $\uparrow$
& 80.00
& 77.00
& 76.75
& 79.50
\\
\midrule

\multirow{3}{*}{GPT-5.2}
& ASR $\downarrow$
& 87.25
& 88.50
& \underline{67.00}
& \textbf{39.75}
\\
& NRP $\uparrow$
& 11.73
& 9.86
& \underline{27.80}
& \textbf{57.09}
\\
& PNA $\uparrow$
& 92.00
& 85.75
& 84.25
& 94.75
\\
\midrule

\multirow{3}{*}{\textbf{Avg.}}
& ASR $\downarrow$
& 85.19
& 83.06
& \underline{63.56}
& \textbf{35.19}
\\
& NRP $\uparrow$
& 9.15
& 9.75
& \underline{20.70}
& \textbf{41.24}
\\
& PNA $\uparrow$
& 61.81
& 57.56
& 56.81
& 63.63
\\

\bottomrule
\end{tabular*}

\endgroup

\caption{
Supplementary ASR, NRP, and PNA results under POT attacks across four LLM backbones.
}
\label{tab:app_pot}
\end{table*}

\begin{table*}[h]
\centering
% \scriptsize
\small
\setlength{\tabcolsep}{3pt}
\renewcommand{\arraystretch}{1.15}
\begin{tabularx}{\textwidth}{
l l
*{5}{>{\centering\arraybackslash}X}
*{5}{>{\centering\arraybackslash}X}
}
\toprule
\multirow{2}{*}{Model} & \multirow{2}{*}{Method}
& \multicolumn{5}{c}{EHRAgent on EICU-AC}
& \multicolumn{5}{c}{SeeAct on Mind2Web-SC} \\
\cmidrule(lr){3-7} \cmidrule(lr){8-12}
& & LPA & LPP & LPR & EA & FRA & LPA & LPP & LPR & EA & FRA \\
\midrule
\multirow{2}{*}{GPT-4o}
& \texttt{GuardAgent}
& 96.20 & 100.00 & 92.86 & 92.86 & \textbf{100.00}
& 54.54 & 56.25 & \textbf{56.25} & \textbf{48.48} & 54.54 \\
& Ours
& \textbf{96.83} & \textbf{100.00} & \textbf{94.12} & \textbf{94.12} & 96.55
& \textbf{60.38} & \textbf{60.87} & 53.85 & 46.15 & \textbf{66.67} \\
\midrule
\multirow{2}{*}{GPT-5.2}
& \texttt{GuardAgent}
& 88.61 & 92.68 & 95.00 & 85.71 & 91.08
& 54.54 & 55.00 & \textbf{68.75} & \textbf{39.39} & 42.42 \\
& Ours
& \textbf{97.47} & \textbf{97.62} & \textbf{97.62} & \textbf{97.62} & \textbf{94.59}
& \textbf{56.60} & \textbf{55.56} & 57.69 & 38.46 & \textbf{55.56} \\
\midrule
\multirow{2}{*}{Avg.}
& \texttt{GuardAgent}
& 92.40 & 96.34 & 93.93 & 89.28 & 95.54
& 54.54 & 55.62 & \textbf{62.50} & \textbf{43.93} & 48.48 \\
& Ours
& \textbf{97.15} & \textbf{98.81} & \textbf{95.87} & \textbf{95.87} & \textbf{95.57}
& \textbf{58.49} & \textbf{58.22} & 55.77 & 42.31 & \textbf{61.12} \\
\bottomrule
\end{tabularx}
\caption{Additional comparison with \texttt{GuardAgent} under its original evaluation protocol. We report results on EHRAgent with EICU-AC and SeeAct with Mind2Web-SC across two LLM backbones. Following \texttt{GuardAgent}, we report label prediction accuracy (LPA$\uparrow$), label prediction precision (LPP$\uparrow$), label prediction recall (LPR$\uparrow$), explanation accuracy (EA$\uparrow$), and final response accuracy (FRA$\uparrow$). Higher values indicate better performance.}
\label{tab:app_guard}
\end{table*}

\begin{table*}[h]
\centering
\setlength{\tabcolsep}{6pt}
\renewcommand{\arraystretch}{1.15}
\begin{tabularx}{\textwidth}{l l *{3}{>{\centering\arraybackslash}X}}
\toprule
Model & Method & DPI DSR$\uparrow$ & IPI DSR$\uparrow$ & FT$\downarrow$ \\
\midrule
\multirow{2}{*}{GPT-4o}
& \texttt{AgentSentinel} & 0.575 & 0.220 & 0.500 \\
& Ours & \textbf{0.780} & \textbf{0.640} & \textbf{0.205} \\
\midrule
\multirow{2}{*}{GPT-5.2}
& \texttt{AgentSentinel} & 0.740 & 0.700 & 0.280 \\
& Ours & \textbf{0.870} & \textbf{0.910} & \textbf{0.052} \\
\midrule
\multirow{2}{*}{Avg.}
& \texttt{AgentSentinel} & 0.658 & 0.460 & 0.390 \\
& Ours & \textbf{0.825} & \textbf{0.775} & \textbf{0.129} \\
\bottomrule
\end{tabularx}
\caption{Additional comparison with \texttt{AgentSentinel} under its original evaluation protocol. Following \texttt{AgentSentinel}, we report defense success rate under direct prompt injection (DPI DSR$\uparrow$) and indirect prompt injection (IPI DSR$\uparrow$), as well as failure rate (FT$\downarrow$), defined as the percentage of failed tasks where the expected attacks fail. Higher DSR indicates stronger defense performance, while lower FT indicates fewer benign-task failures.}
\label{tab:app_agentsentinel}
\end{table*}
\subsection{Additional Comparisons Following Original Guardrail Evaluations}
\label{app:exp_guard}
% As shown in Tab.~\ref{tab:app_guard}, we add comparison experiments on guardrails-like methods \texttt{GuardAgent} \cite{xiang2024guardagent}, which proposes a guardrail agent that protects target agents by verifying their independent behavior safety at each step. Due to the limited open-sourced works and reproducibility, we compare our method to \texttt{GuardAgent} on two LLM backbones with the same test benchmark and metrics as the \texttt{GuardAgent} paper.

% The results show that on average, our method outperforms \texttt{GuardAgent} with higher label prediction accuracy (LPA), Label Prediction Precision (LPP), and final response accuracy (FRA) on both GPT-4o and GPT-5.2.

% The table shows that, on average, our method outperforms \texttt{GuardAgent} across both GPT-4o and GPT-5.2. We consistently improve label prediction accuracy (LPA$\uparrow$) and precision (LPP$\uparrow$) on both EHRAgent and SecAct, indicating more reliable allow/deny decisions with fewer unnecessary denials. Importantly, we also achieve higher final response accuracy (FRA$\uparrow$), suggesting better preservation of benign-task utility.
% Additionally, our method maintains strong recall (LPR$\uparrow$) and explanation accuracy (EA$\uparrow$) with clear gains on EHRAgent, which indicates fewer missed violations and more accurate explanations.

% At the same time, our method remains competitive in recall (LPR) and explanation accuracy (EA): it brings clear gains on EHRAgent, and stays comparable on SecAct while maintaining the stronger overall decision quality and end-task utility.

We further compare TAS with two guardrail-based agent safety methods, \texttt{GuardAgent}~\cite{xiang2024guardagent} and \texttt{AgentSentinel}~\cite{hu2025agentsentinelendtoendrealtimesecurity}. We follow their original evaluation protocols for fair comparisons and report the results as supplementary experiments.

\noindent\textbf{Comparison with \texttt{GuardAgent}.}
As shown in Tab.~\ref{tab:app_guard}, we compare our TAS with \texttt{GuardAgent}, which verifies agent behavior safety at each step. Following the \texttt{GuardAgent} evaluations, we evaluate on EHRAgent with EICU-AC and SeeAct with Mind2Web-SC across two LLM backbones. TAS achieves better average performance than \texttt{GuardAgent}, with higher LPA, LPP, and FRA on both benchmarks, indicating more reliable allow/deny decisions and better benign-task utility. TAS also maintains strong LPR and EA, especially on EHRAgent.

\noindent\textbf{Comparison with \texttt{AgentSentinel}.}
As shown in Tab.~\ref{tab:app_agentsentinel}, we further compare our TAS with \texttt{AgentSentinel} under its original protocol. We report defense success rate under direct and indirect prompt injection (DPI/IPI DSR$\uparrow$) and FT$\downarrow$, which measures the percentage of failed tasks where the expected attacks fail. TAS improves average DPI DSR from $0.658$ to $0.825$ and IPI DSR from $0.460$ to $0.775$, while reducing FT from $0.390$ to $0.129$, showing stronger defense performance and better preservation of benign task execution.

\subsection{Computational Overhead}
\label{app:compute_overhead}

As shown in Tab.~\ref{tab:compute_overhead}, we report the computation overhead of different defense methods relative to \emph{No Defense}. Our method requires 
% approximately $34\%$ 
additional computational time, mainly due to the sampling-based empirical estimation of the tail-risk distribution in \S~{3.4}. We further provide a component-wise breakdown to identify the source of the overhead. The MDP formulation is used for theoretical modeling and does not introduce extra inference latency in practice. The additional runtime mainly comes from sampling-based empirical risk estimation.

% \begin{table}[h]
% \centering
% \setlength{\tabcolsep}{12pt}
% \renewcommand{\arraystretch}{1.15}
% \caption{Computation overhead of different methods (relative to No Defense).}
% \label{tab:compute_overhead}
% \begin{tabular}{lccccccc}
% \toprule
% Methods & No Defense & Deli. & Para. & Instr. & Rewr. & Sand. & Ours \\
% \midrule
% Computation & 1.00$\times$ & 1.14$\times$ & 1.10$\times$ & 0.98$\times$ & 1.12$\times$ & 1.11$\times$ & 1.34$\times$ \\
% \bottomrule
% \end{tabular}
% \end{table}

% \begin{table}[h]
% \centering
% \setlength{\tabcolsep}{4pt}
% \renewcommand{\arraystretch}{1.15}
% \caption{Computation overhead of different methods (relative to No Defense).}
% \label{tab:compute_overhead}
% \begin{tabularx}{\columnwidth}{l c *{7}{>{\centering\arraybackslash}X}}
% \toprule
% Methods & No Defense & Deli. & Para. & Instr. & Rewr. & Sand. & Ours \\
% \midrule
% Computation & 1.00$\times$ & 1.14$\times$ & 1.10$\times$ & 0.98$\times$ & 1.12$\times$ & 1.11$\times$ & 1.34$\times$ \\
% \bottomrule
% \end{tabularx}
% \end{table}

\begin{table*}[h]
\centering
\setlength{\tabcolsep}{4pt}
\renewcommand{\arraystretch}{1.15}
\begin{tabularx}{\textwidth}{l c *{6}{>{\centering\arraybackslash}X}}
\toprule
\multicolumn{8}{l}{\textbf{Overall computation overhead}} \\
\midrule
Methods & No Defense & Deli. & Para. & Instr. & Rewr. & Sand. & Ours \\
Computation & 1.00$\times$ & 1.14$\times$ & 1.10$\times$ & 0.98$\times$ & 1.12$\times$ & 1.11$\times$ & 2.34$\times$ \\
\midrule
\multicolumn{8}{l}{\textbf{Component-wise overhead of TAS}} \\
\midrule
Component & \multicolumn{1}{c}{Vanilla} & \multicolumn{2}{c}{MDP modeling} & \multicolumn{2}{c}{Evaluator calls} & \multicolumn{2}{c}{Sampling} \\
Inference overhead & \multicolumn{1}{c}{1.000$\times$} & \multicolumn{2}{c}{0.000$\times$} & \multicolumn{2}{c}{0.15$\times$} & \multicolumn{2}{c}{1.19$\times$} \\
\bottomrule
\end{tabularx}
\caption{Computation overhead analysis. The upper panel reports the overall computation overhead of different methods relative to No Defense. The lower panel provides a component-wise breakdown of the overhead introduced by TAS.}
\label{tab:compute_overhead}
\end{table*}

% \subsection{Analysis on Tail Risk}
% % \label{4.5}
% \label{app:exp_analysis_tail}

% \tzlconfirmed{To verify the motivation of whether our method actually constrains tail risks in agent workflows, we conduct a tail-risk visualization analysis comparing our method to other baselines.
% Specifically, we run the same workflow with our method and the baselines to record the risk costs at each step using the same risk scoring.
% We aggregate risk costs from each run and plot the tail quantile curve, in which the x-axis is the tail probability (smaller means rarer), and the y-axis is the risk cost threshold. A point $(x, y)$ means that a fraction $x$ of steps have risk costs more than $y$. Therefore, the closer to the upper left in the figure, the larger the tail risk the point has. }
% % We expect baselines to a sharp rise towards the upper left, while our method stays at the lower left.

% \tzlconfirmed{As shown in Fig.~\ref{fig:tail_four_models}, we find the same trend holding across the four LLM backbones, where baseline curves present a sharp rise towards the upper left, while our method consistently stays at the lower left,
% % our curve stays consistently lower than baselines in rare probability (e.g., $x\le10^{-2}$), 
% indicating that our method has substantially smaller risks in the tails.
% In contrast, baseline curves increase steeply near the left end, showing an apparent tail risk.
% The results confirm that our method effectively suppresses tail risk to ensure better safety in the workflow.}

% \input{figures/analysis_tail}

\subsection{Analysis on the Sampling Size $K$}
\label{app:sampling-size}

In TAS, the sampling size $K$ controls the number of candidate outputs used to empirically estimate the step-wise tail-risk distribution in Eq.~(8). A larger $K$ can provide a more reliable estimate of rare high-risk outcomes, but it also introduces additional computational overhead. Therefore, we conduct a sweep over different values of $K$ to analyze the trade-off between safety performance and computational cost.

Specifically, we evaluate $K\!\in\!\{1, 10, 100, 200, 500, 1000\}$ with GPT-4o under the DPI attack while keeping other hyper-parameters fixed. Tab.~\ref{tab:k-sampling} reports the ASR and the relative computational cost normalized by $K=1$. The results show that increasing $K$ consistently improves safety by reducing ASR, confirming that sampling more candidate outcomes helps capture high-risk tail cases more accurately. However, the computational cost increases rapidly when $K$ becomes large. 
% For example, increasing $K$ from $200$ to $500$ only reduces ASR from $40.35$ to $34.72$, while increasing the computational cost from $1.34\times$ to $3.09\times$. Similarly, $K=1000$ further reduces ASR to $31.09$, but requires $7.27\times$ computational cost.
Based on this trade-off, we set $K=200$ in our main experiments. This choice substantially reduces ASR compared with smaller sampling sizes, while keeping the computational overhead moderate. Thus, $K=200$ serves as a well-balanced operating point between safety and efficiency.

% \begin{table}[h]
% \centering
% \setlength{\tabcolsep}{4pt}
% \renewcommand{\arraystretch}{1.15}
% \begin{tabularx}{\columnwidth}{l c *{6}{>{\centering\arraybackslash}X}}
% \toprule
% $K$ & 1 & 10 & 100 & 200 & 500 & 1000 \\
% \midrule
% ASR$\downarrow$ & 68.58 & 65.23 & 55.20 & \textbf{40.35} & 34.72 & 31.09 \\
% % Computational cost$\downarrow$ & 1.00$\times$ & 1.01$\times$ & 1.15$\times$ & \textbf{1.34$\times$} & 3.09$\times$ & 7.27$\times$ \\
% Computational cost$\downarrow$ & 1.00$\times$ & 1.06$\times$ & 1.59$\times$ & \textbf{2.34$\times$} & 6.09$\times$ & 12.27$\times$ \\
% \bottomrule
% \end{tabularx}
% \caption{Analysis of the sampling size $K$. We report ASR and relative computational cost under different values of $K$. Lower ASR and lower computational cost are better.}
% \label{tab:k-sampling}
% \end{table}

\begin{table}[t]
\centering
\small
\setlength{\tabcolsep}{3pt}
\renewcommand{\arraystretch}{1.02}

\begin{tabular*}{\columnwidth}{
    @{\extracolsep{\fill}}
    l
    *{6}{c}
    @{}
}
\toprule
$K$
& 1
& 10
& 100
& \textbf{200}
& 500
& 1000
\\
\midrule

ASR (\%) $\downarrow$
& 68.58
& 65.23
& 55.20
& \textbf{40.35}
& 34.72
& 31.09
\\

Rel. cost ($\times$) $\downarrow$
& 1.00
& 1.06
& 1.59
& \textbf{2.34}
& 6.09
& 12.27
\\

\bottomrule
\end{tabular*}

\caption{
Effect of sampling size $K$ on ASR and relative computational cost.
}
\label{tab:k-sampling}
\end{table}

\subsection{Robustness to Evaluator Misclassifications}
\label{app:evaluator_robustness}
% Since TAS relies on evaluator scores to estimate step-wise safety costs, the reliability of the evaluator may affect the final safety performance. 
To analyze the sensitivity of TAS to evaluator errors, we conduct a robustness analysis by intentionally introducing $10\%$ score misclassifications into the evaluator outputs. Specifically, we randomly perturb $10\%$ of the evaluator risk scores to simulate imperfect risk assessment, and then evaluate the resulting safety and utility performance. We report the averaged ASR and PNA with settings of DPI in Tab.~1.
% Tab.~\ref{tab:main_results}.

As shown in Tab.~\ref{tab:evaluator_robustness}, TAS remains robust under evaluator misclassifications. Compared with the original setting, introducing $10\%$ score misclassifications only increases ASR from $35.98$ to $36.34$, corresponding to about a $1\%$ relative safety drop. This indicates that TAS is not overly sensitive to occasional evaluator errors. We attribute this robustness to the empirical risk estimation in \S~{3.4}, where TAS estimates the risk distribution using multiple sampled candidate outputs rather than relying on a single evaluator score. As a result, isolated score errors are mitigated by distribution-level tail-risk estimation.

\begin{table}[h]
\centering
\setlength{\tabcolsep}{4pt}
\renewcommand{\arraystretch}{1.15}
\begin{tabularx}{\columnwidth}{l >{\centering\arraybackslash}X >{\centering\arraybackslash}X}
\toprule
Method & ASR$\downarrow$ & PNA$\uparrow$ \\
\midrule
% Ours & \textbf{35.98} & \textbf{43.79} \\
% Ours + $10\%$ score misclassifications & 36.34 & 40.86 \\
Ours & \textbf{45.15} & \textbf{63.63} \\
Ours + $10\%$ score misclassifications & 45.61 & 59.37 \\
\bottomrule
\end{tabularx}
\caption{Robustness analysis under evaluator score misclassifications. We intentionally introduce $10\%$ score misclassifications into evaluator outputs.}
\label{tab:evaluator_robustness}
\end{table}

\subsection{False Positive and False Negative Rates}
\label{app:fpr_fnr}

To further evaluate the reliability of risk evaluation, we report both the false positive rate (FPR$\downarrow$) and false negative rate (FNR$\downarrow$) in Tab.~\ref{tab:fpr_fnr}. FPR measures the rate of benign cases incorrectly flagged as risky, while FNR measures the rate of risky cases incorrectly missed by the defense. Lower values indicate better risk evaluation.

The results show that TAS achieves both low FPR and low FNR. Compared with other agent-safety baselines, TAS maintains a substantially lower FPR than all baselines, while also avoiding the high FNR observed in \texttt{Firewalls}, \texttt{DRIFT}, and \texttt{Sentinel}. This indicates that TAS provides a more balanced risk evaluation, reducing unnecessary interventions on benign cases while still detecting risky cases effectively.

\begin{table}[h]
\centering
\setlength{\tabcolsep}{4pt}
\renewcommand{\arraystretch}{1.15}
\begin{tabularx}{\columnwidth}{l *{2}{>{\centering\arraybackslash}X}}
\toprule
Method & FPR$\downarrow$ & FNR$\downarrow$ \\
\midrule
% TAS & {0.057} & 0.015 \\
% {Progent} & 0.235 & {0.000} \\
% {DRIFT} & 0.209 & 0.909 \\
% {Firewalls} & 0.073 & 0.646 \\
% {Sentinel} & 0.105 & 1.000 \\
% {GuardAgent} & 0.871 & 0.097 \\
TAS & {36.68} & 3.93 \\
Default & 43.26 & 47.51 \\
{Progent} & 37.28 & {46.51} \\
{DRIFT} & 69.40 & 15.81 \\
{Firewalls} & 47.21 & 12.74 \\
{Sentinel} & 89.81 & 4.96 \\
{GuardAgent} & 65.07 & 5.04 \\
\bottomrule
\end{tabularx}
\caption{False positive rate and false negative rate of different methods over GPT-4o on AgentDojo. FPR$\downarrow$ measures the rate of benign cases incorrectly flagged as risky, while FNR$\downarrow$ measures the rate of risky cases incorrectly missed. Lower values indicate better risk evaluation.}
\label{tab:fpr_fnr}
\end{table}

\subsection{Significant Test}
\label{app:sig}
As shown in Tab.~\ref{tab:p_asr_simple}, our improvements are statistically significant under the two-sided exact McNemar test with $p<0.01$ on paired task outcomes.

% \begin{table*}[h]
% \centering
% \small
% \setlength{\tabcolsep}{6pt}
% \renewcommand{\arraystretch}{1.1}
% \begin{tabular}{lccccccc}
% \toprule
%  & No Defense & Deli. & Instr. & Para. & Rewr. & Sand. & Shuffle \\
% \midrule
% McNemar $p$ (DPI) 
% & 4.23e-17
% & 7.92e-21 
% & 3.81e-07 
% & 1.46e-03 
% & 1.39e-11 
% & / & /\\
% \midrule
% McNemar $p$ (IPI) 
% & 5.67e-11
% & 4.93e-13 
% & 6.55e-04 
% & / 
% & / 
% & 1.59e-06
% & / \\
% \midrule
% McNemar $p$ (POT)
% & 2.70e-39
% & /
% & /
% & 6.62e-14
% & /
% & /
% & 5.05e-44 \\
% \bottomrule
% \end{tabular}
% \caption{ASR significance test (two-sided exact McNemar).}
% \label{tab:p_asr_simple}
% \end{table*}

\begin{table*}[h]
\centering
\footnotesize
\setlength{\tabcolsep}{5pt}
\renewcommand{\arraystretch}{0.90}

\begingroup
\setlength{\aboverulesep}{0.8pt}
\setlength{\belowrulesep}{0.8pt}

\begin{tabular*}{0.82\textwidth}{
    @{\extracolsep{\fill}}
    lccccccc
    @{}
}
\toprule
&
\textbf{No Defense}
&
\textbf{Deli.}
&
\textbf{Instr.}
&
\textbf{Para.}
&
\textbf{Rewr.}
&
\textbf{Sand.}
&
\textbf{Shuffle}
\\
\midrule

McNemar $p$ (DPI)
& 4.23e-17
& 7.92e-21
& 3.81e-07
& 1.46e-03
& 1.39e-11
& /
& /
\\

McNemar $p$ (IPI)
& 5.67e-11
& 4.93e-13
& 6.55e-04
& /
& /
& 1.59e-06
& /
\\

McNemar $p$ (POT)
& 2.70e-39
& /
& /
& 6.62e-14
& /
& /
& 5.05e-44
\\

\bottomrule
\end{tabular*}

\endgroup

\caption{ASR significance test (two-sided exact McNemar).}
\label{tab:p_asr_simple}
\end{table*}

\section{Details of Step-wise Safety Cost}
\label{app:step_cost}

% (No extra packages needed.) The following boxes use only \fbox + minipage.

\subsection{Rule-based Score}

\noindent
We compute a rule-based score in two stages: (i) a \emph{risk-level pre-classification} from the tool name and arguments, and
(ii) a \emph{feature-based base cost} that scales risk nonlinearly.

\begin{strip}
\centering

% ============================================================
% Left: rule-based pre-classification
% ============================================================
\begin{minipage}[t]{0.51\textwidth}
\begin{riskbox}
\footnotesize
\setlength{\parindent}{0pt}
\setlength{\parskip}{2pt}
\setlength{\emergencystretch}{1em}

\textbf{Rule-based risk-level pre-classification.}
Given a tool name $\texttt{name}$ and arguments $\texttt{args}$,
we flatten $\texttt{args}$ into a string payload and apply:
(a) safe/danger keyword matches,
(b) PII/secret detection, and
(c) external-target checks.

\textbf{Keyword patterns.}
\begin{itemize}[
    leftmargin=1.2em,
    itemsep=0pt,
    topsep=1pt,
    parsep=0pt,
    partopsep=0pt
]
\item \textbf{Safe patterns} (examples):
\texttt{search, fetch, read, retrieve, summarize, translate,
validate, lookup, browse, file\_read}.

\item \textbf{Danger patterns} (examples):
\texttt{send, email, transfer, payment, execute, shell, delete,
publish, upload, credential, bypass, admin, exploit}.
\end{itemize}

\textbf{PII/secret detection.}
We detect PII by regex patterns
(e.g., SSN, credit cards, phone numbers, emails)
and secret keywords
(e.g., \texttt{password, token, private key, credential}),
producing a normalized score
$\mathrm{PII}(\texttt{args})\in[0,1]$.

\textbf{External-target checks.}
We flag an external target if arguments contain
(i) emails not ending with internal domains
(e.g., \texttt{@company.com}), or
(ii) non-internal URLs.

\textbf{Decision rule.}
Let $m_{\text{safe}}$ be the number of safe-pattern matches in
\texttt{name}, and $m_{\text{danger}}$ be the number of danger-pattern
matches in \texttt{name} plus a discounted count from \texttt{args}.
We output the risk level $\ell$ as:
\[
\ell=
\begin{cases}
\texttt{critical},
&
\begin{aligned}[t]
\text{if }&
m_{\text{danger}}\ge 2 \;\text{or}\\
&
\bigl(
m_{\text{danger}}\ge 1
\wedge
(\mathrm{PII}>0.3 \;\text{or external})
\bigr)
\end{aligned}
\\
\texttt{high},
&
\text{if }m_{\text{danger}}\ge 1
\\
\texttt{low},
&
\text{if }m_{\text{safe}}\ge 1
\wedge m_{\text{danger}}=0
\wedge \mathrm{PII}<0.1
\\
\texttt{medium},
&
\text{otherwise.}
\end{cases}
\]
\end{riskbox}
\end{minipage}
\hfill
% ============================================================
% Right: feature-based cost
% ============================================================
\begin{minipage}[t]{0.475\textwidth}
\begin{riskbox}
\footnotesize
\setlength{\parindent}{0pt}
\setlength{\parskip}{2pt}
\setlength{\emergencystretch}{1em}

\textbf{Feature-based base cost.}
We extract interpretable features
(read/write/network/irreversible, external target, PII score,
and tool-name risk), and compute a non-linear base cost:
\[
c_{\text{base}}
=
b(\ell)\,
\phi_{\text{op}}\,
\phi_{\text{ext}}\,
\phi_{\text{pii}}
+
\phi_{\text{name}},
\]
where the terms are:

\begin{itemize}[
    leftmargin=1.2em,
    itemsep=0pt,
    topsep=1pt,
    parsep=0pt,
    partopsep=0pt
]
\item \textbf{Risk-level base:}
$b(\ell)\in\{0.5,2.0,6.0,12.0\}$ for
$\ell\in\{\texttt{low, medium, high, critical}\}$.

\item \textbf{Operation factor:}
$\phi_{\text{op}}=0.8$ for read-only;
$\phi_{\text{op}}=1.3$ for write;
$\phi_{\text{op}}=1.8$ for network/send;
$\phi_{\text{op}}=2.2$ for irreversible actions.

\item \textbf{External factor:}
$\phi_{\text{ext}}
=1+0.8\cdot\mathbb{I}[\text{external target}]$.

\item \textbf{PII factor (nonlinear):}
$\phi_{\text{pii}}
=1+1.5\cdot(\mathrm{PII})^{1.5}$.

\item \textbf{Name penalty:}
$\phi_{\text{name}}=2.0\cdot r_{\text{name}}$,
where $r_{\text{name}}\in[0,1]$ counts high-risk substrings
(e.g., \texttt{transfer, payment, credential, bypass, admin}).
\end{itemize}

\textbf{Rule-only fallback evaluator score
(when the API scorer is disabled).}
When the API scorer is disabled, we use the following normalized
evaluator score only to modulate the stochastic cost distribution:
\[
s_{\text{rule}}(\ell)=
\begin{cases}
0.15, & \ell=\texttt{low},\\
0.40, & \ell=\texttt{medium},\\
0.70, & \ell=\texttt{high},\\
0.90, & \ell=\texttt{critical}.
\end{cases}
\]
\end{riskbox}
\end{minipage}

\end{strip}

\subsection{API-based Score}

\noindent
We use an external LLM API as a risk scorer. The API receives the tool name, a truncated argument string, and the extracted
features, and returns a JSON object with a risk \texttt{level}, \texttt{score}, and \texttt{reason}.

% \begin{center}
% \fbox{
% \begin{minipage}{0.96\textwidth}
% \footnotesize\ttfamily
% \textbf{LLM risk-evaluator prompt (template).}\\[0.2em]
% You are a security risk evaluator for AI agents.\\
% Evaluate the risk of this tool invocation.\\[0.2em]
% Risk Scale:\\
% - 'low' (score 0.0--0.3): Safe operations with minimal potential impact.\\
% - 'medium' (score 0.3--0.6): Moderate risk operations with possible internal or external impact.\\
% - 'high' (score 0.6--0.85): Risky operations with significant potential consequences.\\
% - 'critical' (score 0.85--1.0): Highly risky or dangerous operations with severe impact.\\[0.4em]
% Tool: \{tool\_name\}\\
% Args: \{json\_args\_truncated\}\\
% Features: \{json\_features\}\\[0.3em]
% (Optional) Attack goal: \{tool\_attack\_goal\}\\
% (Optional) Pre-classification suggests: \{risk\_level\}\\[0.4em]
% Return ONLY: \{"level": "low|medium|high|critical", "score": 0.xx, "reason": "..."\}\\
% \end{minipage}}
% \end{center}

\begin{tcolorbox}[
    enhanced,
    breakable,
    width=\linewidth,
    colback=white,
    colframe=black!55,
    boxrule=0.45pt,
    arc=1pt,
    left=5pt,
    right=5pt,
    top=4pt,
    bottom=4pt,
    before skip=3pt,
    after skip=3pt,
    title={LLM risk-evaluator prompt (template).},
    fonttitle=\footnotesize\bfseries\ttfamily,
    fontupper=\footnotesize\ttfamily
]
\raggedright
\setlength{\parindent}{0pt}
\setlength{\parskip}{0pt}
\setlength{\emergencystretch}{1em}

You are a security risk evaluator for AI agents.\par
Evaluate the risk of this tool invocation.\par
\smallskip

Risk Scale:\par
- 'low' (score 0.0--0.3): Safe operations with minimal potential impact.\par
- 'medium' (score 0.3--0.6): Moderate risk operations with possible internal or external impact.\par
- 'high' (score 0.6--0.85): Risky operations with significant potential consequences.\par
- 'critical' (score 0.85--1.0): Highly risky or dangerous operations with severe impact.\par
\smallskip

Tool: \{tool\_name\}\par
Args: \{json\_args\_truncated\}\par
Features: \{json\_features\}\par
\smallskip

(Optional) Attack goal: \{tool\_attack\_goal\}\par
(Optional) Pre-classification suggests: \{risk\_level\}\par
\smallskip

Return ONLY: \{"level": "low|medium|high|critical",
"score": 0.xx, "reason": "..."\}
\end{tcolorbox}

\section{Societal Impacts and Limitations}
\label{app:impacts_limitations}

\subsection{Societal Impacts}
\label{app:societal_impacts}
% This paper proposes Tail-risk-aware Agent Safety (TAS), which models agent workflow safety as a sequence of dependent state transitions and constrains the extreme tail risk using a shared budget. 
% TAS effectively improves the safety in critical cases while preserving utility in normal cases.
% As far as we know, our work has no ethical impacts or potential social consequences.

This paper proposes Tail-risk-aware Agent Safety (TAS), which models agent workflow safety as a sequence of dependent state transitions and constrains the extreme tail risk using a shared budget. 
TAS can have positive societal impacts by improving the safety of LLM-based agents in critical multi-step workflows, reducing rare but severe failures, and helping prevent unsafe tool use or harmful actions while preserving normal task utility.
At the same time, TAS may also introduce potential negative impacts if the risk evaluator is biased or inaccurate. For example, biased evaluator scores may lead to unnecessary refusals of benign tasks and missed risky behaviors.
We further discuss evaluator reliability and utility-safety limitations in App.~\ref{app:limitations}.

\subsection{Limitations}
\label{app:limitations}

% Our TAS has several limitations. First, TAS relies on rule-based and LLM-based evaluators to estimate step-wise safety costs, which may introduce potential biases and compromise risk-estimation accuracy in some cases. While our sampling-based empirical estimation in \S~\ref{3.4} and the analyses in App.~\ref{app:evaluator_robustness} and App.~\ref{app:fpr_fnr} suggest reasonable robustness, designing more unbiased and generalizable risk evaluators remains an important challenge for future work. 
% Second, although TAS achieves better utility than most baselines, it may still slightly compromise benign-task utility and introduce additional computational cost, since stronger safety constraints and larger sampling sizes can reject or modify some otherwise useful executions and require more risk-estimation computation. Users with strict utility or latency requirements may need to carefully tune safety-related hyper-parameters. Nevertheless, TAS remains suitable for safety-critical agent scenarios where reducing high-risk behaviors is prioritized.
% App.~\ref{app:evaluator_robustness} and App.~\ref{app:fpr_fnr}
Our TAS has several limitations. First, TAS relies on rule-based and LLM-based evaluators to estimate step-wise safety costs, which may introduce potential biases and compromise risk-estimation accuracy in some cases. While our sampling-based empirical estimation in \S~{3.4} and the analyses in App.~\ref{app:evaluator_robustness} and App.~\ref{app:fpr_fnr} suggest reasonable robustness, designing more unbiased and generalizable risk evaluators remains an important challenge for future work. Another promising direction is to incorporate richer environmental abstractions and more dynamic risk-budget allocation across agent workflows. 
Second, although TAS achieves better utility than most baselines, it may still slightly compromise benign-task utility and introduce additional computational cost, since stronger safety constraints and larger sampling sizes can reject or modify some otherwise useful executions and require more risk-estimation computation. Users with strict utility or latency requirements may need to carefully tune safety-related hyper-parameters. Nevertheless, TAS remains suitable for safety-critical agent scenarios where reducing high-risk behaviors is prioritized.

% Check whether the conference requires a reproducibility checklist to be included in the paper.
% If so, you can uncomment the following line and ajust the path to include it.
% \input{ReproducibilityChecklist.tex}

\end{document}